\documentclass[]{siamonline1116}

\usepackage[utf8]{inputenc}
\usepackage{graphicx} 
\usepackage{amssymb}
\usepackage{dsfont}
\usepackage{todonotes}
\usepackage[export]{adjustbox}
\usepackage{tikz,pgfplots}
\usetikzlibrary{calc}
\let\oldchardef\chardef
\let\chardef\mathchardef
\usepackage{program}
\let\chardef\oldchardef

\usepackage{color}
\usepackage{microtype}
\usepackage{array,multirow}
\usepackage{rotating}
\usepackage[normalem]{ulem}
\usepackage{booktabs}
\usepackage{graphicx,subcaption}
\usepackage{xcolor}
\usepackage{float}
\usepackage[flushleft]{threeparttable}
\usepackage{lipsum}
\usepackage{amsfonts}
\usepackage{graphicx}
\usepackage{epstopdf}
\usepackage{algorithmic}
\usepackage{xr}
\usepackage{amsopn}
\usepackage{flafter}
\usepackage{placeins} 

\DeclareMathOperator*{\argmin}{argmin}
\DeclareMathOperator{\sign}{sign}
\newcommand{\me}{\mathrm{e}} 
 
\newcommand{\eqdef}{\mathrel{\mathop:}=}

\graphicspath{{figures/}}

\ifpdf
\DeclareGraphicsExtensions{.eps,.pdf,.png,.jpg}
\else
\DeclareGraphicsExtensions{.eps}
\fi

\numberwithin{theorem}{section}

\title{Wasserstein Dictionary Learning:\\ Optimal Transport-Based Unsupervised Nonlinear Dictionary Learning}
\author{Morgan A. Schmitz%
    \thanks{Astrophysics Department, IRFU, CEA, Université Paris-Saclay, F-91191 Gif-sur-Yvette, France
        and
        Université Paris-Diderot, AIM, Sorbonne Paris Cité, CEA, CNRS, F-91191 Gif-sur-Yvette, France (\email{morgan.schmitz@cea.fr})}
    \and 
    Matthieu Heitz%
    \thanks{Université de Lyon, CNRS/LIRIS, Lyon, France}
    \and 
    Nicolas Bonneel%
    \footnotemark[2]
    \and 
    Fred Ngolè%
    \thanks{LIST, Data Analysis Tools Laboratory, CEA Saclay, France}
    \and 
    David Coeurjolly%
    \footnotemark[2]
    \and 
    Marco Cuturi%
    \thanks{Centre de Recherche en Economie et Statistique, Paris, France}
    \and 
    Gabriel Peyré%
    \thanks{DMA, ENS Ulm, Paris, France}
    \and 
    Jean-Luc Starck%
    \footnotemark[1]
}

\headers{Wasserstein Dictionary Learning}
{Schmitz \& al.}

\begin{document}
\maketitle

\begin{abstract}
    This paper introduces a new nonlinear dictionary learning method for histograms in the probability simplex. The method leverages optimal transport theory, in the sense that our aim is to reconstruct histograms using so-called displacement interpolations (a.k.a. Wasserstein barycenters) between dictionary atoms; such atoms are themselves synthetic histograms in the probability simplex. Our method simultaneously estimates such atoms, and, for each datapoint, the vector of weights that can optimally reconstruct it as an optimal transport barycenter of such atoms. Our method is computationally tractable thanks to the addition of an entropic regularization to the usual optimal transportation problem, leading to an approximation scheme that is efficient, parallel and simple to differentiate. Both atoms and weights are learned using a gradient-based descent method. Gradients are obtained by automatic differentiation of the generalized Sinkhorn iterations that yield barycenters with entropic smoothing. Because of its formulation relying on Wasserstein barycenters instead of the usual matrix product between dictionary and codes, our method allows for nonlinear relationships between atoms and the reconstruction of input data. We illustrate its application in several different image processing settings.
\end{abstract}



\section{Introduction}
The idea of dimensionality reduction is as old as data analysis~\cite{pearson1901liii}. Dictionary learning~\cite{lee1999learning}, independent component analysis~\cite{hyvarinen2004}, sparse coding~\cite{lee2007efficient}, autoencoders~\cite{hinton2006reducing} or most simply principal component analysis (PCA) are all variations of the idea that each datapoint of a high-dimensional dataset can be efficiently encoded as a low-dimensional vector. Dimensionality reduction typically exploits a sufficient amount of data to produce an encoding map of datapoints into smaller vectors, coupled with a decoding map able to reconstruct an approximation of the original datapoints using such vectors. Algorithms to carry out the encoding and/or the decoding can rely on simple linear combinations of vectors, as is the case with PCA and nonnegative matrix factorization. They can also be highly nonlinear and employ kernel methods~\cite{scholkopf1997kernel} or neural networks for that purpose~\cite{hinton2006reducing}.

\begin{figure}[h]
    \centering
    \includegraphics[scale=.3]{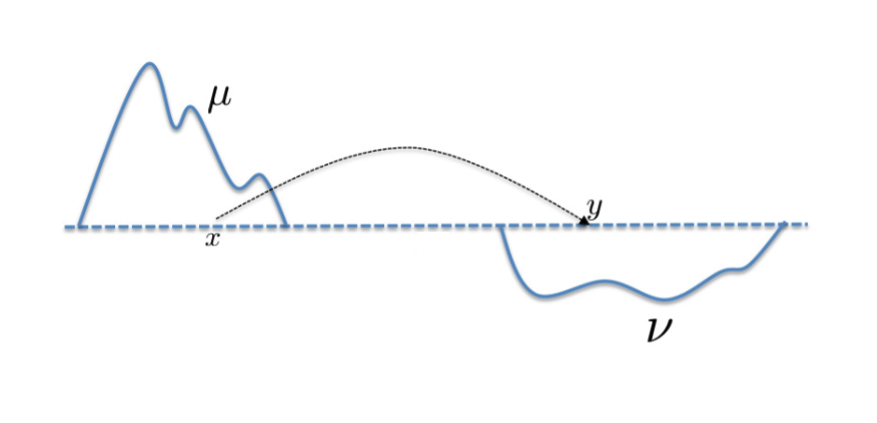}
    \caption{Graphical representation of the mass transportation problem. The minimal effort cost to transport one measure into the other defines the OT distance between $\mu$ and $\nu$.}
    \label{fig:heapsand}
\end{figure}

In this work, we consider a very specific type of encoding/decoding pair, which relies on optimal transport (OT) geometry between probability measures. OT geometry, also known as Wasserstein or earth mover's, defines a distance between two probability measures $\mu,\nu$ by computing the minimal effort required to morph measure $\mu$ into measure $\nu$. Monge's original interpretation~\cite{monge1781} was that $\mu$ would stand for a heap of sand, which should be used to fill in a hole in the ground of the shape of $\nu$. The effort required to move the pile of sand is usually parameterized by a cost function to move one atom of sand from any location $x$ in the support of $\mu$ to any location $y$ in the support of $\nu$ (see \autoref{fig:heapsand}). Monge then considered the problem of finding the optimal (least costly) way to level the ground by transporting the heap into the hole. That cost defines a geometry between probability measures which has several attractive properties. In this paper we exploit the fact that shapes and, more generally, images can be cast as probability measures, and we propose several tools inherited from OT geometry, such as OT barycenters, to warp and average such images~\cite{solomon2015}. These tools can be exploited further to carry out non-linear inverse problems in a Wasserstein sense~\cite{bonneel2016}, and we propose in this work to extend this approach to carry out nonlinear dictionary learning on images using Wasserstein geometry.

\subsection{Previous works}\label{sec:PW}
\subsubsection*{Linear dictionary learning}\label{sec:PWlinear} Several dimensionality reduction approaches rely on using a predefined orthogonal basis upon which datapoints can be projected. Such basis are usually defined without even looking at data, as is the case for Fourier transforms or wavelet-based dictionaries~\cite{mallat1999}. Dictionary learning methods instead underline the idea that dictionaries should be \textit{customized} to fit a particular dataset in an optimal way.
Suppose that the $M$ datapoints of interest can be stored in a matrix $X=(x_1,\dots,x_M)\in\mathbb{R}^{N\times M}$. The aim of (linear) dictionary learning is to factorize the data matrix $X$ using two matrices: a dictionary, $D$, whose elements (the atoms) have the same dimension $N$ as those of $X$, and a list of codes $\Lambda$ used to relate the two: $X \approx D\Lambda$. 

When no constraints on $D$ or $\Lambda$ are given, and one simply seeks to minimize the Frobenius norm of the difference of $X$ and $D\Lambda$, the problem amounts to computing the singular value decomposition of $X$ or, equivalently, the diagonalization of the variance matrix of $X$. In practical situations, one may wish to enforce certain properties of that factorization, which can be done in practice by adding a prior or a constraint on the dictionary $D$, the codes $\Lambda$, or both. For instance, an $l_0$ or $l_1$ norm penalty on the codes yields a sparse representation of data~\cite{aharon2006,mairal2010}. The sparsity constraint might instead be imposed upon the new components (or atoms), as is the case for sparse PCA~\cite{daspremont2007}. Properties other than sparsity might be desired, for example, statistical independence between the components, yielding independent component analysis (ICA~\cite{hyvarinen2004}), or positivity of both the dictionary entries and the codes, yielding a nonnegative matrix factorization (NMF~\cite{lee1999learning}). A third possible modification of the dictionary learning problem is to change the \emph{fitting loss} function that measures the discrepancy between a datapoint and its reconstruction. 
When data lies in the nonnegative orthant, Lee and Seung have shown, for instance, the interest of considering the Kullback-Leibler (KL) divergence to compute such a loss~\cite{lee1999learning} or, more recently, the Wasserstein distance~\cite{rolet2016}, as detailed later in this section. More advanced fitting losses can also be derived using probabilistic graphical models, such as those considered in the topic modeling literature~\cite{blei2009topic}. 

\subsubsection*{Nonlinear dictionary learning}\label{sec:PWnonlinear}
The methods described above are linear in the sense that they attempt to reconstruct each datapoint $x_i$ by a linear combination of a few dictionary elements. Nonlinear dictionary learning techniques involve reconstructing such datapoints using nonlinear operations instead. Autoencoders~\cite{hinton2006reducing} propose using neural networks and to use their versatility to encode datapoints into low-dimensional vectors and later decode them with another network to form a reconstruction. The main motivation behind principal geodesic analysis~\cite{geodesicPCA} is to build such nonlinear operations using geometry, namely by replacing linear interpolations with geodesic interpolations. Of particular relevance to our paper is the body of work that relies on Wasserstein geometry to compute geodesic components \cite{bigot2013,boissard2015distribution,seguy2015,Wang2013} (see \autoref{sec:WPG}).

More generally, when data lies on a Riemannian manifold for which Riemannian exponential and logarithmic maps are known, Ho, Xie and Vemuri propose a generalization of both sparse coding and dictionary learning~\cite{ho2013}. Nonlinear dictionary learning can also be performed by relying on the ``kernel trick'', which allows one to learn dictionary atoms that lie in some feature space of higher, or even infinite, dimension~\cite{harandi2015,liu2015,vannguyen2013}. Equiangular kernel dictionary learning, proposed by Quan, Bao, and Ji, further enforces stability of the learned sparse codes~\cite{quan2016}. Several problems where data is known to belong to a specific manifold are well studied within this framework, \textit{e.g.} sparse coding and dictionary learning for Grassmann manifolds~\cite{harandi2013}, or for positive definite matrices~\cite{harandi2012}, and methods to find appropriate kernels and make full use of the associated manifold's geometry have been proposed for the latter~\cite{li2013}. Kernel dictionary learning has also been studied for the (nonlinear) adaptive filtering framework, where Gao et al. propose an online approach that discards obsolete dictionary elements as new inputs are acquired~\cite{gao2014}. These methods rely on the choice of a particular feature space and an associated kernel and achieve nonlinearity through the use of the latter. The learned dictionary atoms then lie in that feature space. Conversely, our proposed approach requires no choice of kernel. Moreover, the training data and the atoms we learn belong to the same probability simplex, which allows for easy representation and interpretation; \textit{e.g.} our learned atoms can (depending on the chosen fitting loss) capture the extreme states of a transformation undergone by the data. This is opposed to kernel dictionary atoms, which cannot be naturally represented in the same space as datapoints because of their belonging to the chosen high-dimensional feature space.

\subsubsection*{Computational optimal transport}\label{sec:PWcompOT} Optimal transport has seen significant interest from mathematicians in recent decades~\cite{rachev1998mass,talagrand1996transportation,villani2003}. For many years, that theory was, however, of limited practical use and mostly restricted to the comparison of small histograms or point clouds, since typical algorithms used to compute them, such as the auction algorithm~\cite{bertsekas1988} or the Hungarian algorithm~\cite{kuhn1955}, were intractable beyond a few hundred bins or points. Recent approaches~\cite{rabin2011,shirdhonkar2008approximate} have ignited interest for fast yet faithful approximations of OT distances. 
Of particular interest to this work is the entropic regularization scheme proposed by Cuturi~\cite{cuturi2013}, which finds its roots in the gravity model used in transportation theory~\cite{erlander1990gravity}. This regularization can also be tied to the relation between OT and Schrödinger's problem~\cite{schrodinger1931} (as explored by Léonard~\cite{leonard2014}). Whereas the original OT problem is a \emph{linear} problem, regularizing it with an entropic regularization term results in a strictly convex problem with a unique solution which can be solved with Sinkhorn's fixed-point algorithm~\cite{sinkhorn1967}, a.k.a. block coordinate ascent in the dual regularized OT problem. 
That iterative fixed-point scheme yields a numerical approach relying only on elementwise operations on vectors and matrix-vector products. The latter can in many cases be replaced by a separable convolution operator~\cite{solomon2015}, forgoing the need to manipulate a full cost matrix of prohibitive dimensions in some use cases of interest (\textit{e.g.} when input measures are large images).

\subsubsection*{Wasserstein barycenters}\label{sec:PWWassBar} Agueh and Carlier introduced the idea of a Wasserstein barycenter in the space of probability measures~\cite{agueh2011}, namely Fr\'echet means~\cite{frechet1948elements} computed with the Wasserstein metric. Such barycenters are the basic building block of our proposal of a nonlinear dictionary learning scheme with Wasserstein geometry. Agueh and Carlier studied several properties of Wasserstein barycenters and showed very importantly that their exact computation for empirical measures involves solving a multimarginal optimal transport problem, namely a linear program with the size growing exponentially with the size of the support of the considered measures. 

Since that work, several algorithms have been proposed to efficiently compute these barycenters~\cite{bonneel2015,carlier2014numerical,rabin2011,solomon2014wasserstein,yejianbo2017}.
The computation of such barycenters using regularized distances~\cite{cuturi2014} is of particular interest to this work. Cuturi and Peyré~\cite{cuturi2016} use entropic regularization and duality to cast a wide range of problems involving Wasserstein distances (including the computation of Wasserstein barycenters) as simple convex programs with closed form derivatives. They also illustrate the fact that the smoothness introduced by the addition of the entropic penalty can be \textit{desirable}, beyond its computational gains, in the case of the Wasserstein barycenter problem. Indeed, when the discretization grid is small, its true optimum can be highly unstable, which is counteracted by the smoothing introduced by the entropy~\cite[§3.4]{cuturi2016}. The idea of performing iterative Bregman projections to compute approximate Wasserstein distances can be extended to the barycenter problem, allowing its direct computation using a generalized form of the Sinkhorn algorithm~\cite{benamou2015}. Chizat et al. recently proposed a unifying framework for solving unbalanced optimal transport problems~\cite{chizat2016}, including computing a generalization of the Wasserstein barycenter.

\subsubsection*{Wasserstein barycentric coordinates}
An approach to solving the inverse problem associated with Wasserstein barycenters was recently proposed~\cite{bonneel2016}: Given a database of $S$ histograms, a vector of $S$ weights can be associated to any new input histogram, such that the barycenter of that database with those weights approximates as closely as possible the input histogram. These weights are obtained by automatic differentiation (with respect to the weights) of the generalized Sinkhorn algorithm that outputs the approximate Wasserstein barycenter. This step can be seen as an analogy of, given a dictionary $D$, finding the best vector of weights $\Lambda$ that can help reconstruct a new datapoint using the atoms in the dictionary. That work can be seen as a precursor for our proposal, whose aim is to learn both weights \emph{and} dictionary atoms.

\subsubsection*{Applications to image processing}\label{sec:PWAppliImage}
OT was introduced into the computer graphics community by Rubner, Tomasi, and Guibas~\cite{rubner2000} to retrieve images from their color distribution, by considering images as distributions of pixels within a 3-dimensional color space. Color processing has remained a recurring application of OT, for instance to color grade an input image using a photograph of a desired color style~\cite{pitie2005}, or using a database of photographs~\cite{bonneel2016}, or to harmonize multiple images' colors~\cite{bonneel2015}.
Another approach considers grayscale images as 2-dimensional histograms. OT then allows one to find a transport-based warping between images~\cite{haker2004,merigot2011}. 
Further image processing applications are reviewed in the habilitation dissertation of Papadakis~\cite{papadakis2015}. 

\subsubsection*{Wasserstein loss and fidelity}\label{sec:PWWassLoss}
Several recent papers have investigated the use of OT distances as fitting losses that have desirable properties that KL or Euclidean distances cannot offer. We have already mentioned generalizations of PCA to the set of probability measures via the use of OT distances~\cite{bigot2013,seguy2015}. Sandler and Lindenbaum first considered the NMF problem with a Wasserstein loss~\cite{sandler2009}. Their computational approach was, however, of limited practical use. More scalable algorithms for Wasserstein NMF and (linear) dictionary learning were subsequently proposed~\cite{rolet2016}.  
The Wasserstein distance was also used as a loss function with desirable robustness properties to address multilabel supervised learning problems~\cite{frogner2015}. 

Using the Wasserstein distance to quantify the fit between data (an empirical measure) and a parametric family of densities, or a generative model defined using a parameterized push-forward map of a base measure, has also received ample attention in the recent literature. Theoretical properties of such estimators were established by Bassetti, Bodini, and Regazzini~\cite{bassetti2006a} and Bassetti and Regazzini~\cite{bassetti2006b}, and additional results by Bernton et al.~\cite{bernton2017}. Entropic smoothing has facilitated the translation of these ideas into practical algorithms, as illustrated in the work by Montavon, Müller, and Cuturi, who proposed to estimate the parameters of restricted Boltzmann machines using the Wasserstein distance instead of the KL divergence~\cite{montavon2016}. Purely generative models, namely, degenerate probability measures defined as the push-forward of a measure supported on a low-dimensional space into a high-dimensional space using a parameterized function, have also been fitted to observations using a Wasserstein loss~\cite{bernton2017}, allowing for density fitting without having to choose summary statistics (as is often the case with usual methods). The Wasserstein distance has also been used in the context of generative adversarial networks (GANs)~\cite{arjovsky2017}. In that work, the authors use a proxy to approximate the 1-Wasserstein distance. Instead of computing the 1-Wasserstein distance using 1-Lipschitz functions, a classic result from Kantorovich's dual formulation of OT (see Theorem 1.14 in Villani's book~\cite{villani2003}), the authors restrict that set to multilayer networks with rectified linear units and boundedness constraints on weights, which allows them to enforce some form of Lipschitzness of their networks. Unlike the entropic smoothing used in this paper, that approximation requires solving a nonconvex problem whose optimum, even if attained, would be arbitrarily far from the true Wassertein distance. More recently, Genevay, Peyré, and Cuturi introduced a general scheme for using OT distances as the loss in generative models~\cite{genevay2017}, which relies on both the entropic penalty and automatic differentiation of the Sinkhorn algorithm. Our work shares some similarities with that paper, since we also propose automatically differentiating the Sinkhorn iterations used in Wasserstein barycenter computations.

\subsection{Contributions}
In this paper, we introduce a new method for carrying out nonlinear dictionary learning for probability histograms using OT geometry. Nonlinearity comes from the fact that we replace the usual linear combination of dictionary atoms by Wasserstein barycenters. Our goal is to reconstruct datapoints using the closest (according to any arbitrary fitting loss on the simplex, not necessarily the Wasserstein distance) Wasserstein barycenter to that point using the dictionary atoms. 
Namely, instead of considering linear reconstructions for $X \approx D\Lambda$, our aim is to approximate columns of $X \approx \mathbf{P}(D,\Lambda)$ using the $\mathbf{P}$ operator which maps atoms $D$ with lists of weights $\Lambda$ to their respective barycenters.

Similar to many traditional dictionary learning approaches, this is achieved by finding local minima of a nonconvex energy function. To do so, we propose using automatic differentiation of the iterative scheme used to compute Wasserstein barycenters. We can thus obtain gradients with respect to both the dictionary atoms and the weights that can then be used within one's solver of choice (in this work, we chose to use an off-the-shelf quasi-Newton approach and perform both dictionary and code updates simultaneously).

Our nonlinear dictionary learning approach makes full use of the Wasserstein space's properties, as illustrated in \autoref{fig:gaussex}: two atoms are learned from a dataset made up of five discretized Gaussian distributions in 1D, each slightly translated on the grid. Despite the simplicity of the transformation (translation), linear generative models fail to capture the changes of the geometrical space, as opposed to our OT approach. Moreover, the atoms we learn are also discrete measures, unlike the PCA and NMF components.

We also offer several variants and improvements to our method:
\begin{itemize}
    \item Arbitrarily sharp reconstructions can be reached by performing the barycenter computation in the log-domain;
    \item We offer a general method to make use of the separability of the kernel involved and greatly alleviate the computational cost of this log-domain stabilization;
    \item Our representation is learned from the differentiation of an iterative, Sinkhorn-like algorithm, whose convergence can be accelerated by using information from previous Sinkhorn loops at each initialization (warm start), or adding a momentum term to the Sinkhorn iterations (heavyball);
    \item We expand our method to the unbalanced transport framework.
\end{itemize}

Part of this work was previously presented as a conference proceedings~\cite{schmitz2017}, featuring an initial version of our method, without any of the above improvements and variants, and in the case where we were only interested in learning two different atoms.

Additional background on OT is given in \autoref{sec:OT}. The method itself and an efficient implementation are presented in \autoref{sec:WDL}. We highlight other extensions in \autoref{sec:extensions}. We showcase its use in several image processing applications in \autoref{sec:apps}.

\newcolumntype{R}[1]{>{\centering\let\newline\\\arraybackslash\begin{sideways}}p{#1}<{\end{sideways}}}

\newcommand{\morgancolsize}{0.1\linewidth}
\newcommand{\morganspacer}{\hspace{0.1mm}}

\begin{figure*}
    \centering
    \begin{tabular}{@{}R{0.05\linewidth}@{\morganspacer}>{\centering\let\newline\\\arraybackslash}p{\morgancolsize}@{\morganspacer}>{\centering\let\newline\\\arraybackslash}p{\morgancolsize}@{\morganspacer}>{\centering\let\newline\\\arraybackslash}p{\morgancolsize}@{\morganspacer}>{\centering\let\newline\\\arraybackslash}p{\morgancolsize}@{\morganspacer}>{\centering\let\newline\\\arraybackslash}p{\morgancolsize}@{\morganspacer}>{\centering\let\newline\\\arraybackslash}p{\morgancolsize}@{\morganspacer}>{\centering\let\newline\\\arraybackslash}p{\morgancolsize}}
        (a) Data & & \includegraphics[width=\linewidth]{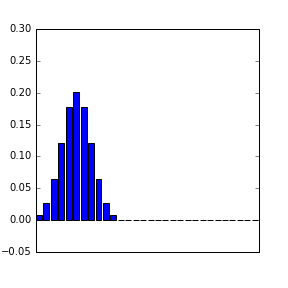} &
        \includegraphics[width=\linewidth]{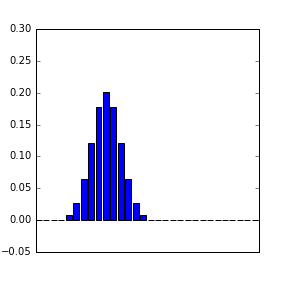} &
        \includegraphics[width=\linewidth]{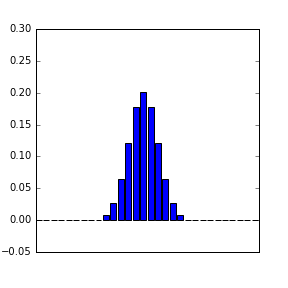} &
        \includegraphics[width=\linewidth]{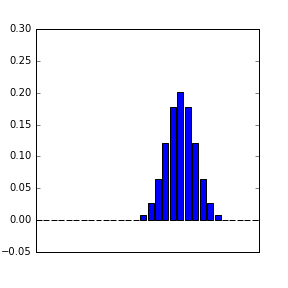} &
        \includegraphics[width=\linewidth]{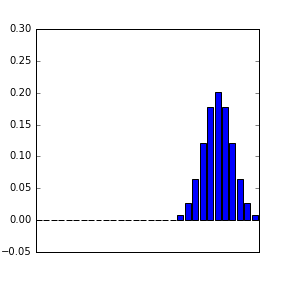} &\\
        
        (b) PCA & \includegraphics[width=\linewidth]{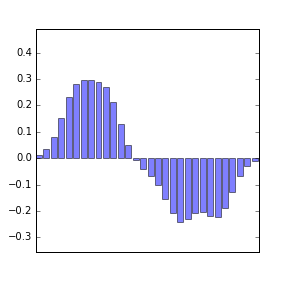} &
        \includegraphics[width=\linewidth]{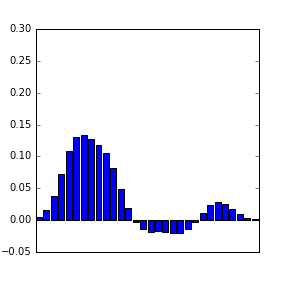} &
        \includegraphics[width=\linewidth]{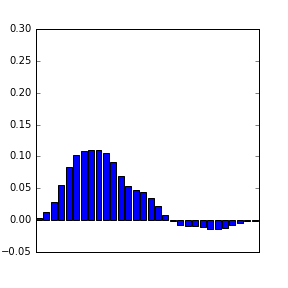} &
        \includegraphics[width=\linewidth]{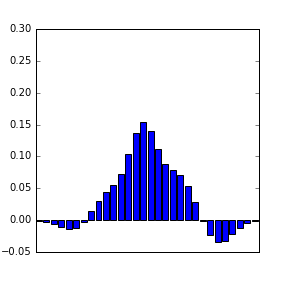} &
        \includegraphics[width=\linewidth]{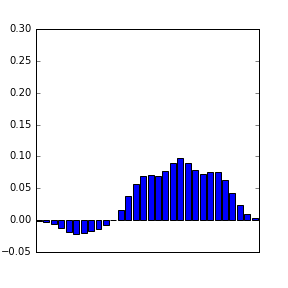} &
        \includegraphics[width=\linewidth]{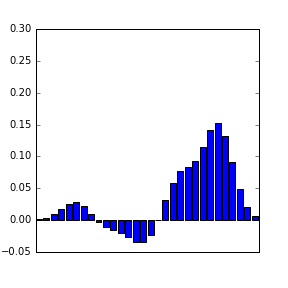} &
        \includegraphics[width=\linewidth]{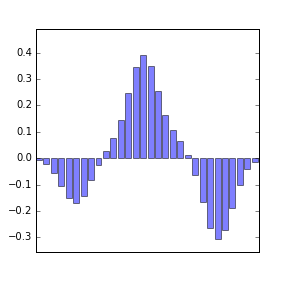}\\
        
        (c) NMF & \includegraphics[width=\linewidth]{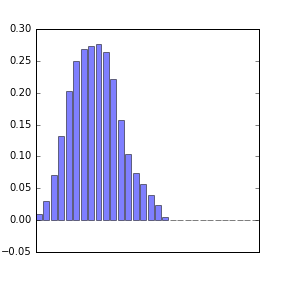} &
        \includegraphics[width=\linewidth]{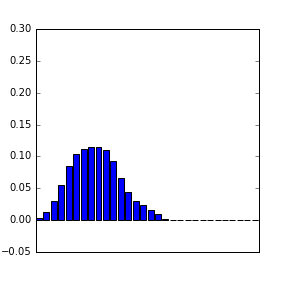} &
        \includegraphics[width=\linewidth]{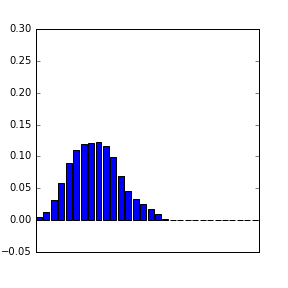} &
        \includegraphics[width=\linewidth]{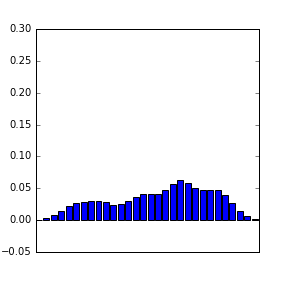} &
        \includegraphics[width=\linewidth]{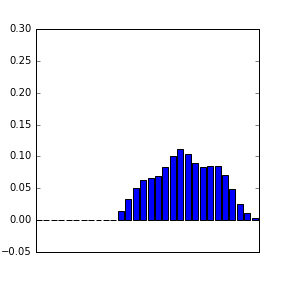} &
        \includegraphics[width=\linewidth]{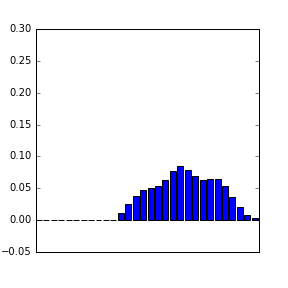} &
        \includegraphics[width=\linewidth]{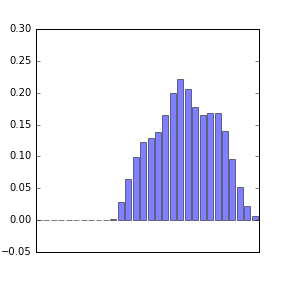}\\
        
        (d) WDL & \includegraphics[width=\linewidth]{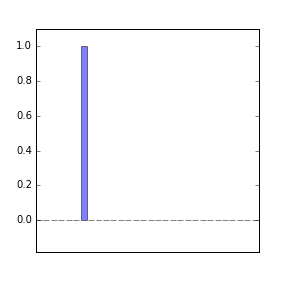} &
        \includegraphics[width=\linewidth]{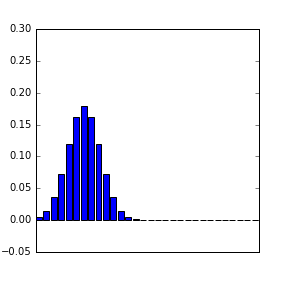} &
        \includegraphics[width=\linewidth]{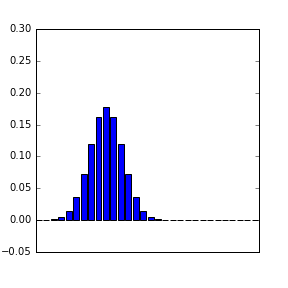} &
        \includegraphics[width=\linewidth]{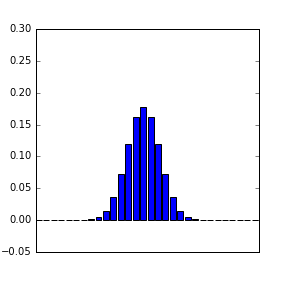} &
        \includegraphics[width=\linewidth]{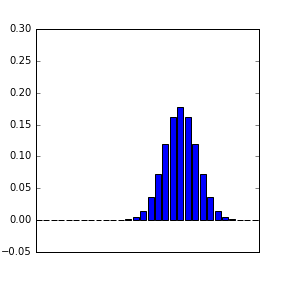} &
        \includegraphics[width=\linewidth]{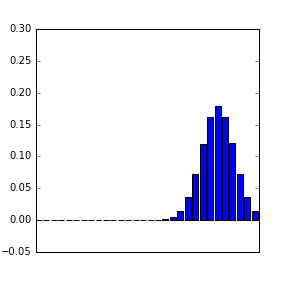} &
        \includegraphics[width=\linewidth]{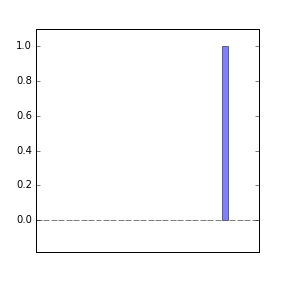}\\
        
    \end{tabular}	
    \caption{Top row: data points. Bottom three rows: On the far sides, in purple, are the two atoms learned by PCA, NMF and our method (WDL), respectively. In between the two atoms are the reconstructions of the five datapoints for each method. The latter two were relaunched a few times with randomized initializations and the best local minimum was kept. As discussed in \autoref{sec:OT}, the addition of an entropy penalty to the usual OT program causes a blur in the reconstructions. When the parameter associated with the entropy is high, our method yields atoms that are sharper than the dataset on which it was trained, as is observed here where the atoms are Dirac despite the dataset consisting of discretized Gaussians. See \autoref{sec:logstabilizations} for a method to reach arbitrarily low values of the entropy parameter and counteract the blurring effect.}
    \label{fig:gaussex}
\end{figure*}

\subsection{Notations}
We denote $\Sigma_d$ the simplex of $\mathbb{R}^d$, that is,

\begin{align*}
\Sigma_d \eqdef \left\{u\in\mathbb{R}^d_+, \sum_{i=1}^du_i = 1\right\}.
\end{align*}
For any positive matrix $T$, we define its negative entropy as

\begin{align*}
H(T) \eqdef \sum_{i,j} T_{ij} \log(T_{ij}-1).
\end{align*}
$\odot$ denotes the Hadamard product between matrices or vectors. Throughout this paper, when applied to matrices, $\prod, \div$, and $\exp$ notations refer to elementwise operators. The scalar product between two matrices denotes the usual inner product, that is,

\begin{align*}
\langle A, B\rangle \eqdef \mathrm{Tr}(A^\top B) = \sum_{i,j}A_{ij}B_{ij},
\end{align*}
where $A^\top$ is the transpose of $A$. For $(p,q)\in\Sigma_N^2$, we denote their set of couplings as

\begin{align}
\label{eq:transportplans}\Pi(p,q) \eqdef \left\{T\in\mathbb{R}^{N\times N}_+, T\mathds{1}_N = p, T^\top\mathds{1}_N=q\right\},
\end{align}
where $\mathds{1}_N = (1,\dots,1)^\top\in\mathbb{R}^N$. $\Delta$ denotes the $\mathrm{diag}$ operator, such that if $u\in\mathbb{R}^N$, then

\begin{align*}
\Delta(u) \eqdef \begin{pmatrix}
u_1 & &\\
& \ddots &\\
& & u_N
\end{pmatrix}\in\mathbb{R}^{N\times N}.
\end{align*}
$\iota$ is the indicator function, such that for two vectors $u,v$,
\begin{align}\label{eq:indicator}
\iota_{\left\{u\right\}}(v) = 
\begin{cases}
0&\mbox{ if } u=v,\\
+\infty&\mbox{ otherwise,}
\end{cases}
\end{align}
and $\mathrm{KL}(.\vert .)$ is their KL divergence, defined here as
\begin{align*}
\mathrm{KL}(u\vert v) = \sum_i u_i \log\left(\frac{u_i}{v_i}\right) -u_i +v_i.
\end{align*}
For a concatenated family of vectors $t = \left[t_1^{\top},\dots,t_S^{\top}\right]^{\top} \in\mathbb{R}^{NS}$, we write the  $i$th element of $t_s$ as $[t_s]_i$.
We denote the rows of matrix $M$ as $M_{i.}$ and its columns as $M_{.j}$. $I_N$ and $\mathbf{0}_{N\times N}$ are the $N\times N$ identity and zero matrices, respectively.

\section{Optimal transport}\label{sec:OT}
\subsection{OT distances}\label{sec:OTdist}
In the present work, we restrict ourselves to the discrete setting, \textit{i.e.} our measures of interest will be histograms, discretized on a fixed grid of size $N$ (Eulerian discretization), and represented as vectors in $\Sigma_N$. In this case, the cost function is represented as a cost matrix $C\in\mathbb{R}^{N\times N}$, containing the costs of transportation between any two locations in the discretization grid. The OT distance between two histograms $(p,q)\in\Sigma_N^2$ is the solution to the discretized Monge--Kantorovich problem:

\begin{align*}
W(p,q) \eqdef \min_{T\in\Pi(p,q)} \langle T,C \rangle.
\end{align*}
As defined in \eqref{eq:transportplans}, $\Pi(p,q)$ is the set of admissible couplings between $p$ and $q$, that is, the set of matrices with rows summing to $p$ and columns to $q$. A solution, $T^*\in\mathbb{R}^{N\times N}$, is an optimal transport plan. 

Villani's books give extended theoretical overviews of OT~\cite{villani2003,villani2008} and, in particular, several properties of such distances. The particular case where the cost matrix is derived from a metric on the chosen discretization grid yields the so-called Wasserstein distance (sometimes called the earth mover's distance). For example, if $C_{ij} = \|x_i-x_j\|_2^2$ (where $x_i,x_j$ are the positions on the grid), the above formulation yields the squared 2-Wasserstein distance, the square-root of which is indeed a distance in the mathematical sense. Despite its intuitive formulation, the computation of Wasserstein distances grows supercubicly in $N$, making them impractical as dimensions reach the order of one thousand grid points. This issue has motivated the recent introduction of several approximations that can be obtained at a lower computational cost (see \hyperref[sec:PWcompOT]{Computational optimal transport, subsection 1.1}). Among such approximations, the entropic regularization of OT distances~\cite{cuturi2013} relies on the addition of a penalty term as follows:

\begin{align}\label{eq:entpen}
W_\gamma(p,q) \eqdef \min_{T\in\Pi(p,q)} \langle T, C\rangle + \gamma H(T),
\end{align}
where $\gamma>0$ is a hyperparameter. As $\gamma\rightarrow 0$, $W_\gamma$ converges to the original Wasserstein distance, while higher values of $\gamma$ promote more diffuse transport matrices. The addition of a negentropic penalty makes the problem $\gamma$-strongly convex; first-order conditions show that the problem can be analyzed as a matrix-scaling problem which can be solved using Sinkhorn's algorithm~\cite{sinkhorn1967} (also known as the iterative proportional fitting procedure (IPFP)~\cite{deming1940}). The Sinkhorn algorithm can be interpreted in several ways: for instance, it can be thought of as an alternate projection scheme under a KL divergence for couplings~\cite{benamou2015} or as a block-coordinate ascent on a dual problem~\cite{cuturi2014}. The Sinkhorn algorithm consists in using the following iterations for $l\geq 1$, starting with $b^{(0)} = \mathds{1}_N$:

\begin{align}
\label{eq:simpleSinkhorn}
a^{(l)} &= \frac{q}{K^\top b^{(l-1)}},\\
\nonumber b^{(l)} &= \frac{p}{Ka^{(l)}},
\end{align}
where $K \eqdef \mathrm{exp}(\frac{-C}{\gamma})$ is the elementwise exponential of the negative of the rescaled cost matrix. Note that when $\gamma$ gets close to $0$, some values of $K$ become negligible, and values within the scaling vectors, $a^{(l)}$ and $b^{(l)}$, can also result in numerical instability in practice (we will study workarounds for that issue in \autoref{sec:logstabilizations}). Application of the matrix $K$ can often be closely approximated by a separable operation~\cite{solomon2015} (see \autoref{sec:logkernel} for separability even in the log-domain). In the case where the histograms are defined on a uniform grid and the cost matrix is the squared Euclidean distance, the convolution kernel is simply Gaussian with standard deviation $\sqrt{\gamma/2}$. The two vectors $a^{(l)}, b^{(l)}$ converge linearly towards the optimal scalings~\cite{franklin1989} corresponding to the optimal solution of \eqref{eq:entpen}. Notice finally that the Sinkhorn algorithm at each iteration $l\geq 1$ results in an approximate optimal transport matrix $T^{(l)} = \Delta(b^{(l)})K\Delta(a^{(l)})$.

\subsection{Wasserstein barycenter}\label{sec:wassbary}
\begin{figure}
    \centering
    \begin{subfigure}{0.49\linewidth}
        \centering
        \includegraphics[width=.72\textwidth]{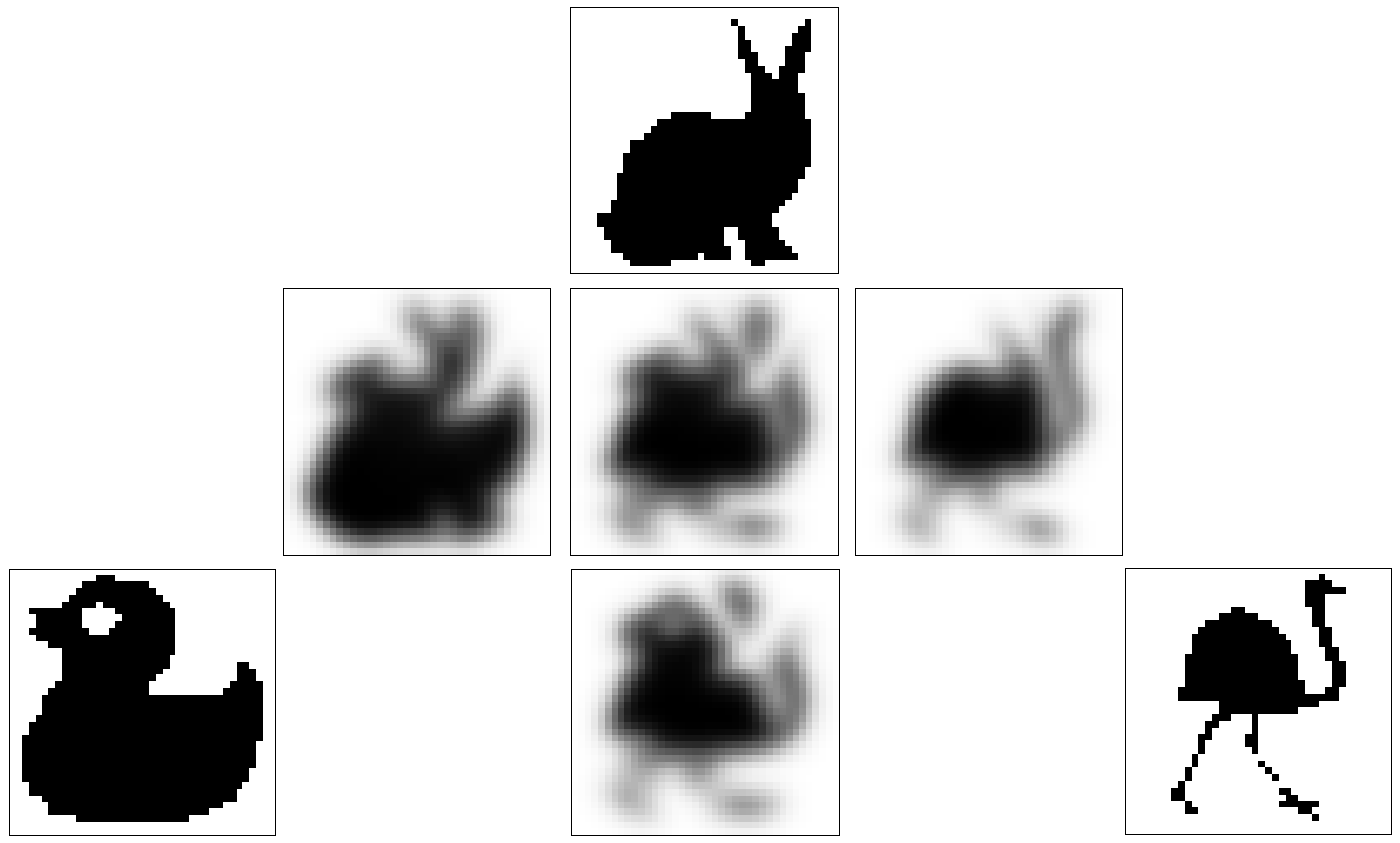}
        \caption{Wasserstein simplex; $\gamma=8$}
    \end{subfigure}
    \begin{subfigure}{0.49\linewidth}
        \centering
        \includegraphics[width=.72\textwidth]{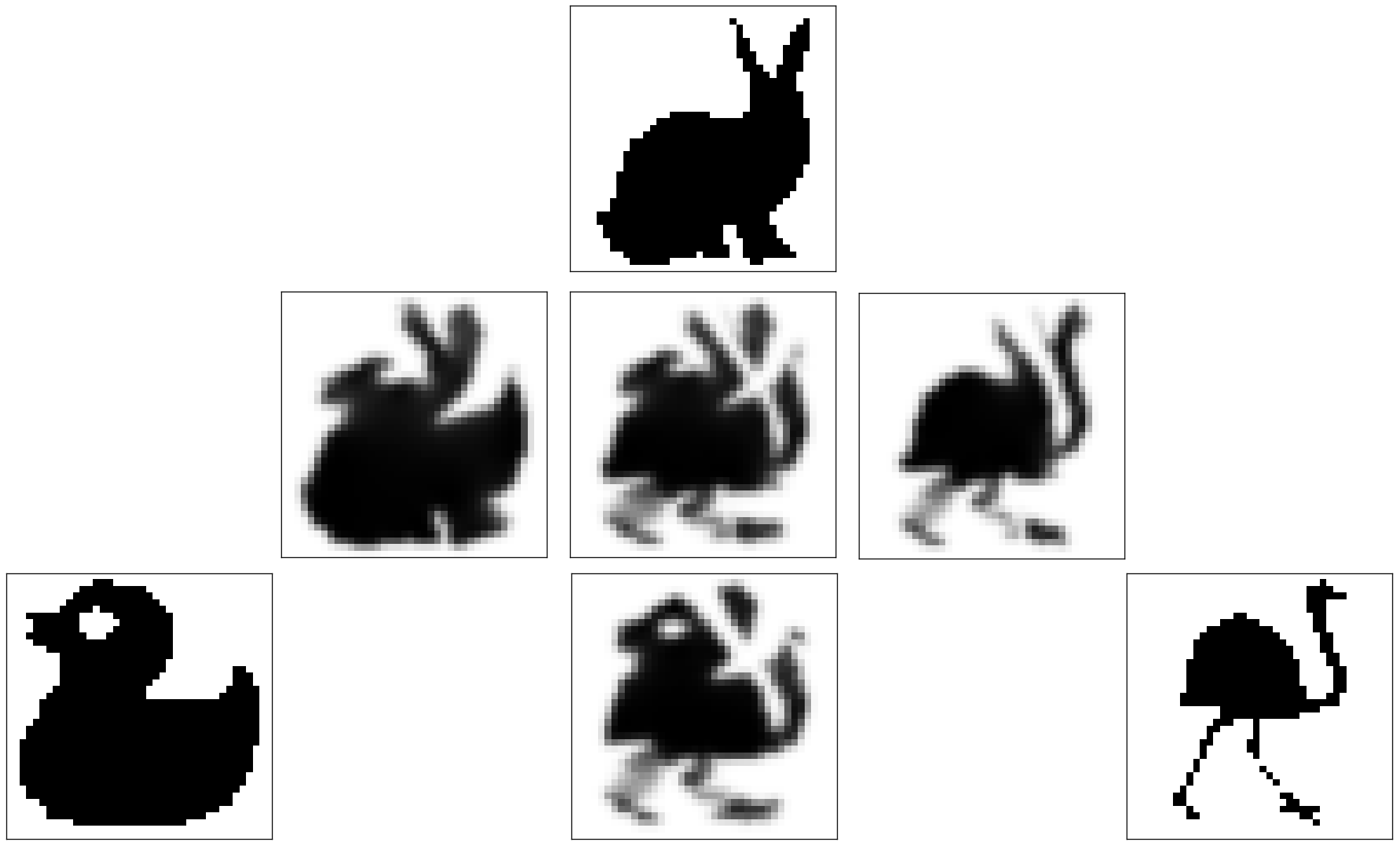}
        \caption{Wasserstein simplex; $\gamma=1$}
    \end{subfigure}
    \caption{Wasserstein simplices: barycenters of the three images in the corners with varying barycentric weights. Middle row: $\lambda = \left[\frac{1}{2},\frac{1}{2}, 0\right], \left[\frac{1}{3},\frac{1}{3},\frac{1}{3}\right], \left[0,\frac{1}{2},\frac{1}{2}\right]$. Bottom row, center: $\left[\frac{1}{2},0,\frac{1}{2}\right]$.}
    \label{fig:simplexes}
\end{figure}

Analogous to the usual Euclidean barycenter, the Wasserstein barycenter of a family of measures is defined as the minimizer of the (weighted) sum of squared Wasserstein distances from the variable to each of the measures in that family~\cite{agueh2011}. For measures with the same discrete support, we define, using entropic regularization, the barycenter of histograms $(d_1,\dots,d_S)\in(\Sigma_N)^S$ with barycentric weights $\lambda = (\lambda_1,\dots,\lambda_S)\in\Sigma_S$ as

\begin{align}\label{eq:defwassbar}
P\left(D,\lambda\right) \eqdef \argmin_{u\in\Sigma_N} \sum_{s=1}^S \lambda_s W_\gamma(d_s,u),
\end{align}
where $D \eqdef (d_1^\top,\dots,d_S^\top)^\top\in\mathbb{R}^{NS}$. The addition of the entropy term ensures strict convexity and thus that the Wasserstein barycenter is uniquely defined. It also yields a simple and efficient iterative scheme to compute approximate Wasserstein barycenters, which can be seen as a particular case of the unbalanced OT setting~\cite{chizat2016}. This scheme, a generalization of the Sinkhorn algorithm, once again relies on two scaling vectors:

\begin{align}
\label{eq:a}a_s^{(l)} &= \frac{d_s}{Kb_s^{(l-1)}},\\
\label{eq:sinkhornbary}P^{(l)}\left(D,\lambda\right) &= \prod_{s=1}^S\left(K^\top a_s^{(l)}\right)^{\lambda_s},  \\
\label{eq:b}b_s^{(l)} &= \frac{P^{(l)}\left(D, \lambda\right)}{K^\top a_s^{(l)}},
\end{align}
where, as before, $K = \mathrm{exp}(\frac{-C}{\gamma})$. In this case, however, the scaling vectors are of size $NS$, such that $a^{(l)}=\left(a_1^{(l)\top},\dots,a_S^{(l)\top}\right)^\top, b^{(l)}=\left(b_1^{(l)\top},\dots,b_S^{(l)\top}\right)^\top$ and $b^{(0)} = \mathds{1}_{NS}$. Note that one can perform both scaling vector updates at once (and avoid storing both) by plugging one of \eqref{eq:a}, \eqref{eq:b} into the other. An illustration of the Wasserstein barycenter, as well as the impact of the $\gamma$ parameter, is given in \autoref{fig:simplexes}.

\section{Wasserstein dictionary learning}\label{sec:WDL}
\subsection{Overview}
Given data $X\in\mathbb{R}^{N\times M}$ in the form of histograms, \textit{i.e.}, each column $x_i\in\Sigma_N$ (for instance a list of $M$ images with normalized pixel intensities), and the desired number of atoms $S$, we aim to learn a dictionary $D$ made up of histograms $(d_1,\dots,d_S)\in\left(\Sigma_N\right)^S$ and a list of barycentric weights $\Lambda = (\lambda_1,\dots,\lambda_M)\in\left(\Sigma_S\right)^M$ so that for each input, $P(D,\lambda_i)$ is the best approximation of $x_i$ according to some criterion $\mathcal{L}$ (see \autoref{tab:losses} for examples). Namely, our representation is obtained by solving the problem

\begin{align}\label{eq:energy}
\min_{D,\Lambda} \mathcal{E}(D,\Lambda) \eqdef \sum_{i=1}^M \mathcal{L}\left(P(D,\lambda_i), x_i\right).
\end{align}
Note the similarity between the usual dictionary learning formulation (see \hyperref[sec:PWlinear]{Linear dictionary learning, subsection 1.1}) and the one above. In our case, however, the reconstruction of the original data happens \textit{via} the nonlinear Wasserstein barycenter operator, $\mathbf{P}(D,\Lambda) = \left(P(D,\lambda_i)\right)_i$, instead of the (linear) matrix product $D\Lambda$.

Differentiation of \eqref{eq:energy} relies in part on the computation of the Wasserstein barycenter operator's Jacobians with regard to either the barycentric weights or the atoms. While it is possible to obtain their analytical formulae, for example by using the fact that Sinkhorn updates \eqref{eq:sinkhornbary}-\eqref{eq:b} become fixed-point equations when convergence is reached, they rely on solving a linear system of prohibitive dimensionality for our settings of interest where $N$ is typically large (Bonneel, Peyré, and Cuturi derived the expression with regard to barycentric weights and discussed the issue in~\cite[§4.1]{bonneel2016}). Moreover, in practice, the true Wasserstein barycenters with entropic penalty $P(D,\lambda_i)$ are unknown and approximated by sufficient Sinkhorn iterations \eqref{eq:sinkhornbary}--\eqref{eq:b}. As is now common practice in some machine learning methods (a typical example being backward propagation for neural nets), and following recent works~\cite{bonneel2016}, we instead take an approach in the vein of automatic differentiation~\cite{griewank2008}. That is, we recursively differentiate the iterative scheme yielding our algorithm instead of the analytical formula of our Wasserstein barycenter. In our case, this is the generalization of the Sinkhorn algorithm for barycenters. Instead of \eqref{eq:energy}, we thus aim to minimize

\begin{align}
\label{eq:energyL}\min_{D,\Lambda} \mathcal{E}_L(D,\Lambda) \eqdef   \sum_{i=1}^M\mathcal{L}\left(P^{(L)}(D,\lambda_i), x_i\right),
\end{align}
where $P^{(L)}$ is the approximate barycenter after $L$ iterations, defined as in \eqref{eq:sinkhornbary}. Even when using an entropy penalty term, we have no guarantee on the convexity of the above problem, whether jointly in $D$ and $\Lambda$ or for each separately, contrary to the case of OT distance computation in \eqref{eq:entpen}. We thus aim to reach a local minimum of energy landscape $\mathcal{E}_L$ by computing its gradients and applying a descent method. By additivity of $\mathcal{E}_L$ and without loss of generality, we will focus on the derivations of such gradients for a single datapoint $x\in\Sigma_N$ (in which case $\Lambda$ only comprises one list of weights $\lambda\in\Sigma_S$). Differentiation of \eqref{eq:energyL} yields

\begin{align}
\label{eq:diffU}\nabla_{D}\mathcal{E}_L(D,\Lambda) &=  \left[\partial_{D}P^{(L)}(D,\lambda)\right]^\top \nabla \mathcal{L}(P^{(L)}(D,\lambda), x),\\
\label{eq:difflbda}\nabla_{\lambda}\mathcal{E}_L(D,\Lambda) &= \left[\partial_{\lambda}P^{(L)}(D,\lambda)\right]^\top \nabla \mathcal{L}(P^{(L)}(D,\lambda), x).
\end{align}
The right-hand term in both cases is the gradient of the loss which is typically readily computable (see \autoref{tab:losses}) and depends on the choice of fitting loss. The left-hand terms are the Jacobians of the Wasserstein barycenter operator with regard to either the weights or the dictionary. These can be obtained either by performing the analytical differentiation of the $P^{(l)}$ operator, as is done in \autoref{sec:backwardrec} (and \autoref{appdx:proof}), or by using an automatic differentiation library such as Theano~\cite{theano2016}. The latter approach ensures that the complexity of the backward loop is the same as that of the forward, but it can lead to memory problems due to the storing of all objects being part of the gradient computation graph (as can be the case, for instance, when performing the forward Sinkhorn loop in the log-domain as in \autoref{sec:stabheuristic}; for this specific case, an alternative is given in \autoref{sec:logkernel}). The resulting numerical scheme relies only on elementwise operations and on the application of the matrix $K$ (or its transpose) which often amounts to applying a separable convolution~\cite{solomon2015} (see \autoref{sec:logkernel}). The resulting algorithm is given in \autoref{alg:grads}. At first, a ``forward'' loop is performed, which amounts to the exact same operations as those used to compute the approximate Wasserstein barycenter using updates \eqref{eq:sinkhornbary}-\eqref{eq:b} (the barycenter for current weights and atoms is thus computed as a by-product). Two additional vectors of size $SNL$ are stored and then used in the recursive backward differentiation loops that compute the gradients with regard to the dictionary and the weights.

\begin{table}[th]
	\begin{threeparttable}
		\centering
		\begin{tabular}{c|c|c}
			Name & $\mathcal{L}(p,q)$ & $\nabla\mathcal{L}$\\
			\hline
			Total variation & $\|p-q\|_1$ & $\sign(p-q)$ \\
			Quadratic & $\|p-q\|^2_2$ & $2(p-q)$\\
			Kullback-Leibler & $\mathrm{KL}(p\vert q)$ & $\log(p/q)-1$\\
			Wasserstein\footnotemark & $W^{(L)}_\gamma(p,q)$ & $\gamma\log(a^{(L)})$\\
		\end{tabular}
	\end{threeparttable}
	\centering
	\caption{Examples of similarity criteria and their gradient in $p$. See \autoref{fig:facesLossesB} for the atoms yielded by our method for these various fitting losses.}
	\label{tab:losses}
\end{table}
\footnotetext{In this case, the loss is computed iteratively as explained in \autoref{sec:OTdist}, and $a^{(L)}$ in the gradient's expression is obtained after $L$ iterations as in \eqref{eq:simpleSinkhorn}.} 

\begin{algorithm}
    \begin{program}
        \mbox{\textbf{Inputs:} Data $x\in\Sigma_N$, atoms $d_1,\dots,d_S\in\Sigma_N$, current weights $\lambda\in\Sigma_S$}
        \color{blue}{\COMMENT{Sinkhorn loop}}
        \forall s, b_s^{(0)}  \eqdef \mathbf{1}_{N}\\
        \FOR l=1 \TO L \STEP 1 \DO
        \forall s, \varphi_s^{(l)} \eqdef K^\top\frac{d_s}{Kb_s^{(l-1)}}\\
        p \eqdef \prod_s \left(\varphi_s^{(l)}\right)^{\lambda_s}\\
        \forall s, b_s^{(l)} \eqdef \frac{p}{\varphi_s^{(l)}}
        \OD
        \color{blue}{\COMMENT{Backward loop - weights}}
        w \eqdef \mathbf{0}_S\\
        r \eqdef \mathbf{0}_{S\times N}\\
        g \eqdef \nabla \mathcal{L}(p,x) \odot p\\
        \FOR l=L \TO 1 \STEP -1 \DO
        \forall s, w_s \eqdef w_s + \langle\log\varphi_s^{(l)}, g \rangle\\
        \forall s, r_s \eqdef - K^\top\left(K\left(\frac{\lambda_s g - r_s}{\varphi_s^{(l)}}\right) \odot \frac{d_s}{(Kb_s^{(l-1)})^2}\right) \odot b_s^{(l-1)}\\
        g \eqdef \sum_s r_s
        \OD
        \color{blue}{\COMMENT{Backward loop - dictionary}}
        y \eqdef \mathbf{0}_{S\times N}\\
        z \eqdef \mathbf{0}_{S\times N}\\
        n \eqdef \nabla \mathcal{L}(p,x)
        \FOR l=L \TO 1 \STEP -1 \DO   
        \forall s, c_s \eqdef K((\lambda_sn - z_s)\odot b_s^{(l)})\\
        \forall s, y_s \eqdef y_s + \frac{c_s}{Kb_s^{(l-1)}}\\
        \forall s, z_s \eqdef -\frac{\mathds{1}_N}{\varphi_s^{(l-1)}}\odot K^\top\frac{d_s\odot c_s}{(Kb_s^{(l-1)})^2}\\
        n \eqdef \sum_s z_s
        \OD
        \mbox{\textbf{Outputs:} $P^{(L)}(D,\lambda) \eqdef p, \nabla_D\mathcal{E}^{(L)} \eqdef y, \nabla_\lambda\mathcal{E}^{(L)} \eqdef w$}
    \end{program}
    \caption{\texttt{SinkhornGrads}: Computation of dictionary and barycentric weights gradients}
    \label{alg:grads}
\end{algorithm}

Using the above scheme to compute gradients, or its automatically computed counterpart from an automatic differentiation tool, one can find a local minimum of the energy landscape \eqref{eq:energyL}, and thus the eventual representation $\Lambda$ and dictionary $D$, by applying any appropriate optimization method under the constraints that both the atoms and the weights belong to their respective simplices $\Sigma_N,\Sigma_S$. 

For the applications shown in \autoref{sec:apps}, we chose to enforce these constraints through the following change of variables

\begin{align*}
&\forall i, d_i \eqdef F_N(\alpha_i) \eqdef \frac{\me^{\alpha_i}}{\sum_{j=1}^N \me^{\left[ \alpha_i\right]_j}},
&\lambda \eqdef F_S(\beta) \eqdef \frac{\me^{\beta}}{\sum_{j=1}^S \me^{\beta_j}}.
\end{align*}
The energy to minimize (with regard to $\alpha, \beta$) then reads as

\begin{align}\label{eq:logisticenergy}
\mathcal{G}_L(\alpha, \beta) \eqdef \mathcal{E}_L(F(\alpha),F_S(\beta)),
\end{align}
where $F(\alpha) \eqdef \left(F_N(\alpha_1),\dots,F_N(\alpha_S)\right) = D$. Differentiating \eqref{eq:logisticenergy} yields

\begin{align*}
\nabla_\alpha\mathcal{G}_L(\alpha,\beta) &= \left[\partial F(\alpha)\right]^\top \nabla_D\mathcal{E}_L\left(F(\alpha),F_S(\beta)\right) = \left[\partial F(\alpha)\right]^\top \nabla_D\mathcal{E}_L\left(D,\Lambda\right),\\
\nabla_\beta\mathcal{G}_L(\alpha,\beta) &= \left[\partial F_S(\beta)\right]^\top \nabla_\lambda\mathcal{E}_L\left(F(\alpha),F_S(\beta)\right) = \left[\partial F_S(\beta)\right]^\top \nabla_\lambda\mathcal{E}_L\left(D,\Lambda\right),
\end{align*}
where $\left[\partial F_p(u)\right]^\top = \partial F_p(u) = \left(I_p - F_p(u)\mathds{1}^\top_p\right)\Delta\left(F_p(u)\right)$, $p$ being either $N$ or $S$ for each atom or the weights, respectively, and both derivatives of $\mathcal{E}_L$ are computed using either automatic differentiation or as given in \eqref{eq:diffU}, \eqref{eq:difflbda} with \autoref{alg:grads} (see \autoref{sec:backwardrec}). The optimization can then be performed with no constraints over $\alpha, \beta$. 

Since the resulting problem is one where the function to minimize is differentiable and we are left with no constraints, in this work we chose to use a quasi-Newton method (though our approach can be used with any appropriate solver); that is, at each iteration $t$, an approximation of the inverse Hessian matrix of the objective function, $B^{(t)}$, is updated, and the logistic variables for the atoms and weights are updated as
\begin{align*}
&\alpha^{(t+1)} \eqdef \alpha^{(t)} - \rho_\alpha^{(t)}B_\alpha^{(t)}\nabla_\alpha\mathcal{G}_L(\alpha,\beta),
&\beta^{(t+1)} \eqdef \beta^{(t)} - \rho_\beta^{(t)}B_\beta^{(t)}\nabla_\beta\mathcal{G}_L(\alpha,\beta),
\end{align*}
where the $\rho^{(t)}$ are step sizes. An overall algorithm yielding our representation in this particular setup of quasi-Newton after a logistic change of variables is given in \autoref{alg:WDL}.

In the applications of \autoref{sec:apps}, $B^{(t)}$ and $\rho^{(t)}$ were chosen using an off-the-shelf L-BFGS solver~\cite{morales2011}. We chose to perform updates to atoms and weights simultaneously. Note that in this case, both are fed to the solver of choice as a concatenated vector. It is then beneficial to add a ``variable scale'' hyperparameter 
$\zeta$ and to multiply all gradient entries related to the weights by that value. Otherwise, the solver might reach its convergence criterion when approaching a local minimum with regards to either dictionary atoms or weights, even if convergence is not yet achieved in the other. Setting either a low or high value of $\zeta$ bypasses the problem by forcing the solver to keep optimizing with regard to one of the two variables in particular. In practice, and as expected, we have observed that relaunching the optimization with different $\zeta$ values upon convergence can increase the quality of the learned representation. While analogous to the usual alternated optimization scheme often used in dictionary learning problems, this approach avoids having to compute two different forward Sinkhorn loops to obtain the derivatives in both variables.

\begin{algorithm}
    \begin{program}
        \mbox{\textbf{Inputs:} Data $X\in\mathbb{R}^{N\times M}$, initial guesses $\alpha^{(0)}, \beta_1^{(0)},\dots,\beta_M^{(0)}$, convergence criterion}
        t\eqdef 0
        \WHILE \mbox{convergence not achieved} \DO
        D^{(t)} \eqdef F(\alpha^{(t)})
        \alpha^{(t+1)} \eqdef \alpha^{(t)}
        \FOR i=1 \TO M \STEP 1 \DO
        \lambda_i^{(t)} \eqdef F_S(\beta_i^{(t)})
        p_i, g^D_i, g^\lambda_i \eqdef \texttt{SinkhornGrads}(x_i, D^{(t)}, \lambda_i^{(t)})
        \mbox{Select } \rho_\alpha^{(t)},\rho_i^{(t)} B_\alpha^{(t)}, B_i^{(t)} \mbox{ (\verb|L-BFGS|)}
        \alpha^{(t+1)} \eqdef \alpha^{(t+1)} - \rho_\alpha^{(t)}B_\alpha^{(t)}\partial F(\alpha^{(t)})g^D_i 
        \beta_i^{(t+1)} \eqdef \beta_i^{(t)} - \rho_i^{(t)}B_i^{(t)}\partial F_S(\beta_i^{(t)})g^{\lambda}_i 
        \OD
        t \eqdef t+1
        \OD
        \mbox{\textbf{Outputs:} $D=F\left(\alpha^{(t)}\right)$, $\Lambda=\left(F_S\left(\beta_1^{(t)}\right), \dots, F_S\left(\beta_S^{(t)}\right)\right)$}
    \end{program}
    \caption{Quasi-Newton implementation of the Wasserstein dictionary learning algorithm}
    \label{alg:WDL}
\end{algorithm}

\subsection{Backward recursive differentiation}
\label{sec:backwardrec}
To differentiate $P^{(L)}(D,\Lambda)$, we first rewrite its definition \eqref{eq:sinkhornbary} by introducing the following notations:

\begin{align}
\label{eq:updateP}P^{(l)}(D,\lambda) &= \Psi(b^{(l-1)}(D,\lambda), D, \lambda),\\
\label{eq:updateb}b^{(l)}(D,\lambda) &= \Phi(b^{(l-1)}(D,\lambda),D,\lambda),
\end{align}
where

\begin{align}
\label{eq:defPsi}
\Psi(b,D,\lambda) &\eqdef \prod_s \left(K^\top\frac{d_s}{Kb_s}\right)^{\lambda_s},\\
\label{eq:defPhi}\Phi(b,D,\lambda) &\eqdef \left[\left(\frac{\Psi(b,D,\lambda)}{K^\top\frac{d_1}{Kb_1}}\right)^\top,\dots,\left(\frac{\Psi(b,D,\lambda)}{K^\top\frac{d_S}{Kb_S}}\right)^\top\right]^\top.
\end{align}
Finally, we introduce the following notations for readability:

\begin{align*}
&\xi_y^{(l)} \eqdef \left[\partial_y\xi(b^{(l)},D,\lambda)\right]^\top,
&B_y^{(l)} \eqdef \left[\partial_yb^{(l)}(D,\lambda)\right]^\top,
\end{align*}
where $\xi$ can be $\Psi$ or $\Phi$, and $y$ can be $D$ or $\lambda$.

\begin{proposition}\label{prop:jacks}
    \begin{align}
    \label{eq:nabUE}\nabla_D \mathcal{E}_L(D,\lambda)= \Psi_D^{(L-1)}\left(\nabla\mathcal{L}(P^{(L)}(D,\lambda),x)\right) + \sum_{l=0}^{L-2} \Phi_D^{(l)}\left(v^{(l+1)}\right),\\
    \label{eq:nablbdaE}\nabla_\lambda \mathcal{E}_L(D,\lambda)= \Psi_\lambda^{(L-1)}\left(\nabla\mathcal{L}(P^{(L)}(D,\lambda),x)\right) + \sum_{l=0}^{L-2} \Phi_\lambda^{(l)}\left(v^{(l+1)}\right),
    \end{align}
\end{proposition}
where:

\begin{align}
\label{eq:vL}
v^{(L-1)} &\eqdef \Psi_b^{(L-1)} \left(\nabla L(P^{(L)}(D,\lambda),x) \right),\\
\label{eq:vlm1}
\forall l < L-1, v^{(l-1)} &\eqdef \Phi_b^{(l-1)}\left(v^{(l)}\right).
\end{align}

See \autoref{appdx:proof}  proof.

\section{Extensions}\label{sec:extensions}
\subsection{Log-domain stabilization} \label{sec:logstabilizations}
\subsubsection{Stabilization}\label{sec:stabheuristic}
In its most general framework, representation learning aims at finding a useful representation of data, rather than one allowing for perfect reconstruction. In some particular cases, however, it might also be desirable to achieve a very low reconstruction error, for instance if the representation is to be used for compression of data rather than a task such as classification. In the case of our method, the quality of the reconstruction is directly linked to the selected value of the entropy parameter $\gamma$, as it introduces a blur in the reconstructed images as illustrated in \autoref{fig:simplexes}. In the case where sharp features in the reconstructed images are desired, we need to take extremely low values of $\gamma$, which can lead to numerical problems, \textit{e.g.} because values within the scaling vectors $a$ and $b$ can then tend to infinity. As suggested by Chizat et al.~\cite{chizat2016} and Schmitzer~\cite{schmitzer2016}, we can instead perform the generalized Sinkhorn updates \eqref{eq:sinkhornbary}-\eqref{eq:b} in the log-domain. Indeed, noting $u_s^{(l)}, v_s^{(l)}$ as the dual scaling variables, that is,

\begin{align*}
&a_s^{(l)} \eqdef \exp\left(\frac{u_s^{(l)}}{\gamma}\right),
&b_s^{(l)} \eqdef \exp\left(\frac{v_s^{(l)}}{\gamma}\right),
\end{align*}
the quantity $-c_{ij} + u_i + v_j$ is known to be bounded and thus remains numerically stable. We can then introduce the stabilized kernel $\tilde{K}(u,v)$ defined as

\begin{align}
\label{eq:stabilizedkernel}
\tilde{K}(u, v) \eqdef \exp\left(\frac{-C + u\mathds{1}^\top + \mathds{1}v^\top}{\gamma}\right),
\end{align}
and notice that we then have

\begin{align*}
u_s^{(l)} &= \gamma \left[\log (d_s) - \log(Kb_s^{(l-1)})\right],\\
\left[\log(Kb_s^{(l-1)})\right]_i &= \log\left[\sum_j\exp\left(\frac{-c_{ij}+v_j^{(l-1)}}{\gamma}\right)\right]\\
&= \log\left(\sum_j \tilde{K}(u^{(l-1)}_s,v^{(l-1)}_s)_{.j}\right) - \frac{\left[u_s^{(l-1)}\right]_i}{\gamma}.
\end{align*}
With similar computations for the $v_s$ updates, we can then reformulate the Sinkhorn updates in the stabilized domain as

\begin{align}
\label{eq:alphaupdate}u_s^{(l)} \eqdef \gamma\left[\log(d_s) - \log\left(\sum_j \tilde{K}(u^{(l-1)}_s,v^{(l-1)}_s)_{.j}\right)\right] + u_s^{(l-1)},\\
\label{eq:betaupdate}v_s^{(l)} \eqdef \gamma\left[\log(P^{(l)}) - \log\left(\sum_i \tilde{K}(u_s^{(l)},v_s^{(l-1)})_{i.}\right)\right] + v_s^{(l-1)}.
\end{align}
This provides a forward scheme for computing Wasserstein barycenters with arbitrarily low values of $\gamma$, which could be expanded to the backward loop of our method either by applying an automatic differentiation tool to the stabilized forward barycenter algorithm or by changing the steps in the backward loop of \autoref{alg:grads} to make them rely solely on stable quantities. However, this would imply computing a great number of stabilized kernels as in \eqref{eq:stabilizedkernel}, which relies on nonseparable operations. Each of those kernels would also have to either be stored in memory or recomputed when performing the backward loop. In both cases, the cost in memory or number of operations, respectively, can easily be too high in large scale settings.

\subsubsection{Separable log kernel}\label{sec:logkernel}

These issues can be avoided by noticing that when the application of the kernel $K$ is separable, this operation can be performed at a much lower cost.
For a $d$-dimensional histogram of $N=n^d$ bins, applying a separable kernel amounts to performing a sequence of $d$ steps, where each step computes $n$ operations per bin.
It results in a $O(n^{d+1}) = O(N^{\frac{d+1}{d}})$ cost instead of $O(N^2)$.
As mentioned previously, the stabilized kernel \eqref{eq:stabilizedkernel} is not separable, prompting us to introduce a new stable and separable kernel suitable for log-domain processing.
We illustrate this process using 2-dimensional kernels without loss of generality.
Let $\mathcal{X}$ be a 2-dimensional domain discretized as an $n\times n$ grid.
Applying a kernel of the form  $K=\exp(-\frac{C}{\gamma})$ to  a 2-dimensional image $b \in \mathcal{X}$ is performed as such:
\begin{align*}
R(i,j) \eqdef \sum^n_{k=1} \sum^n_{l=1} \exp\left(-\frac{C((i,j),(k,l))}{\gamma}\right) b(k,l)\,,
\end{align*}
where $C((i,j),(k,l))$ denotes the cost to transport mass between the points $(i,j)$ and $(k,l)$.

Assuming a separable cost such that ${C((i,j),(k,l)) \eqdef C_y(i,k) + C_x(j,l)}$\,, it amounts to performing two sets of 1-dimensional kernel applications:
\begin{align*}
A(k,j) &= \sum^n_{l=1} \exp\left(\frac{C_x(j,l)}{\gamma}\right) b(k,l), \\
R(i,j) &= \sum^n_{k=1} \exp\left(\frac{C_y(i,k)}{\gamma}\right) A(k,j)\,.
\end{align*}

In order to stabilize the computation and avoid reaching representation limits, we transfer it to the log-domain ($v \eqdef \log(b)$).
Moreover, we shift the input values by their maximum and add it at the end.
The final process can be written as the operator $K_{LS} : \log(b) \rightarrow \log(K(b))$ with $K$ a separable kernel, and is described in \autoref{alg:logSepKer}.

\begin{algorithm}
    \begin{program}
        \mbox{\textbf{Inputs:} Cost matrix $C\in\mathbb{R}^{N\times N}$, image in log-domain $v \in \mathbb{R}^{n\times n}$}
        
        \forall k,j,~ x_l(k,j) \eqdef \frac{C_x(j,l)}{\gamma} + v(k,l) \\ 
        \forall k,j,~ A'(k,j) \eqdef \log\left(\sum^n_l \exp(x_l - \max_l x_l)\right) + \max_l x_l \\ 
        \forall i,j,~ y_k(i,j) \eqdef \frac{C_y(i,k)}{\gamma} + A'(k,j) \\
        \forall i,j,~ R'(i,j) \eqdef \log\left(\sum^n_k \exp(y_k - \max_k y_k)\right) + \max_k y_k \\ 
        \mbox{\textbf{Outputs:} Image in log-domain $K_{LS}(v)=R'$}
    \end{program}
    \caption{\texttt{LogSeparableKer} $K_{LS}$: Application of a 2-dimensional separable kernel in log-domain}
    \label{alg:logSepKer}
\end{algorithm}

This operator can be used directly in the forward loop, as seen in \autoref{alg:logGrads}.
For backward loops, intermediate values can be negative and real-valued logarithms are not suited. While complex-valued logarithms solve this problem, they come at a prohibitive computational cost.
Instead, we store the sign of the input values and compute logarithms of absolute values.
When exponentiating, the stored sign is used to recover the correct value.

\subsection{Warm start}
Warm start, often used in optimization problems, consists in using the solution of a previous optimization problem, close to the current one, as the initialization point in order to speed up the convergence. Our method relies on performing an iterative optimization process (for example, L-BFGS in the following experiments) which, at each iteration, calls upon \textit{another} iterative scheme: the forward Sinkhorn loop to compute the barycenter and its automatic differentiation to obtain gradients. As described in \autoref{sec:wassbary}, this second, nested iterative scheme is usually initialized with constant scaling vectors. However, in our case, since each iteration of our descent method performs a new Sinkhorn loop, the scaling vectors of the previous iteration can be used to set the values of $b^{(0)}$ instead of the usual $\mathds{1}_{NS}$, thus ``warm-starting'' the barycenter computation. In the remainder of this subsection, for illustrative purposes, we will focus on our particular case where the chosen descent method is L-BFGS, though the idea of applying warm start to the generalized Sinkhorn algorithm should be directly applicable with any other optimization scheme.

As an example, in our case, instead of a single L-BFGS step after $L=500$ Sinkhorn iterations, we perform an L-BFGS step every $L=10$ iterations, initializing the scaling vectors as the ones reached at the end of the previous 10.
This technique accumulates the Sinkhorn iterations as we accumulate L-BFGS steps. This has several consequences: a gain in precision and time, a potential increase in the instability of the scaling vectors, and changes in the energy we minimize.

First, the last scaling vectors of the previous overall iteration are closer to that of the current one than a vector of constant value.
Therefore, the Sinkhorn algorithm converges more rapidly, and the final barycenters computed at each iteration gain accuracy compared to the classical version of the algorithm.

Second, as mentioned in \autoref{sec:logstabilizations}, the scaling vectors may become unstable when computing a large number of iterations of the Sinkhorn algorithm.
When using a warm start strategy, Sinkhorn iterations tend to accumulate, which may consequently degrade the stability of the scaling vectors.
For example, using 20 Sinkhorn iterations running through 50 L-BFGS steps, a warm start would lead to barycenters computed using scaling vectors comparable to those obtained after 1000 Sinkhorn iterations.
When instabilities become an issue, we couple the warm start approach with our log-domain stabilization.
The reduced speed of log-domain computations is largely compensated by the fact that our warm start allows the computation of fewer Sinkhorn iterations for an equivalent or better result.

Third, when differentiating \eqref{eq:energyL}, we consider the initial, warm-started (as opposed to initializing $b^{(0)}$ to $\mathds{1}_{NS}$) values given to the scaling vectors to be constant and independent of weights and atoms. This amounts to considering a different energy to minimize at each L-BFGS step.

\begin{figure}
    \centering
    \begin{subfigure}{0.48\linewidth}
        \begin{tikzpicture}
        \begin{axis}[
        xlabel={Number of L-BFGS iterations},
        ylabel={Mean PSNR},
        width=0.95\textwidth,
        height=5cm,
        xmin = 0, xmax = 500,
        ymin = 32, ymax = 34,
        legend pos = south east
        ]
        
        \addplot[
        color=green!60!black,
        mark=dot,
        mark size=1pt
        ]
        table [x=Iterations, y=Normal-Niter2, col sep=comma] {graphs/J0329-mug_001_expr2-KLloss.csv};
        
        \addplot[
        color=blue,
        mark=dot,
        mark size=1pt
        ]
        table [x=Iterations, y=Normal-Niter100, col sep=comma] {graphs/J0329-mug_001_expr2-KLloss.csv};
        
        \addplot[
        color=red,
        mark=circle,
        mark size=1pt
        ]
        table [x=Iterations, y=WR10-LR10-Niter2, col sep=comma] {graphs/J0329-mug_001_expr2-KLloss.csv};
        
        \legend{$L=2$ (20m), $L=100$ (15h), \shortstack{Warm start \& \\ $L=2$ (25m)}}
        \end{axis}
        \end{tikzpicture}
        \caption{\textbf{MUG dataset: woman}}
        \label{fig:warmstart001}
    \end{subfigure}
    \begin{subfigure}{0.48\linewidth}
        \begin{tikzpicture}
        \begin{axis}[
        xlabel={Number of L-BFGS iterations},
        ylabel={Mean PSNR},
        width=0.95\textwidth,
        height=5cm,
        xmin = 0, xmax = 500,
        ymin = 30.8, ymax = 32.1,
        legend pos = south east
        ]
        
        \addplot[
        color=green!60!black,
        mark=dot,
        mark size=1pt
        ]
        table [x=Iterations, y=Normal-Niter2, col sep=comma] {graphs/J0373-mug_014_expr2-KLloss.csv};
        
        \addplot[
        color=blue,
        mark=dot,
        mark size=1pt
        ]
        table [x=Iterations, y=Normal-Niter100, col sep=comma] {graphs/J0373-mug_014_expr2-KLloss.csv};
        
        \addplot[
        color=red,
        mark=circle,
        mark size=1pt
        ]
        table [x=Iterations, y=WR10-LR10-Niter2, col sep=comma] {graphs/J0373-mug_014_expr2-KLloss.csv};
        
        \legend{$L=2$ (20m), $L=100$ (15h), \shortstack{Warm start \& \\ $L=2$ (25m)}}
        \end{axis}
        \end{tikzpicture}
        \caption{\textbf{MUG dataset: man}}
        \label{fig:warmstart014}
    \end{subfigure}
    \caption{Evolution of the mean PSNR of the reconstructions per L-BFGS iteration, for three configurations, on two datasets. The KL loss was used for this experiment. We see that the warm start yields better reconstructions with the same number of Sinkhorn iterations ($L$) in roughly the same time.}
    \label{fig:warmstart}
\end{figure}
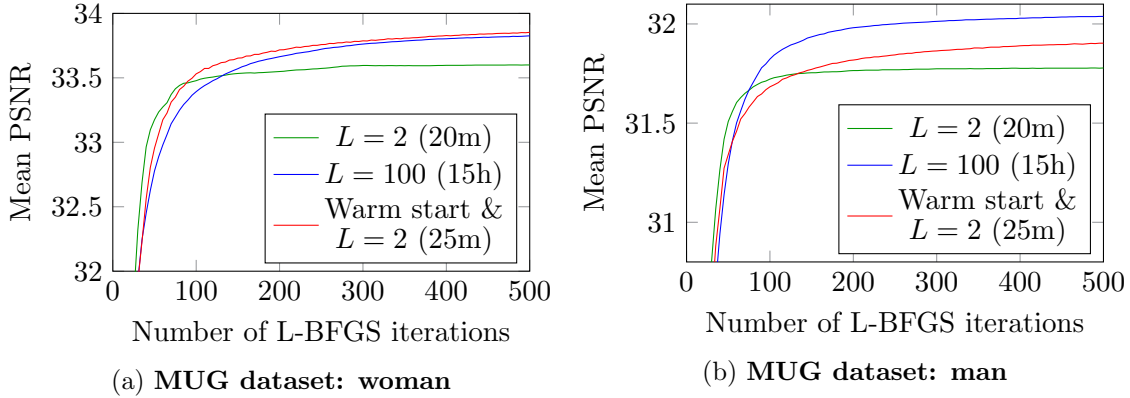

We demonstrate the benefits of the warm start in \autoref{fig:warmstart}.
We plot the evolution of the mean peak signal-to-noise ratio (PSNR) of the reconstructions throughout the L-BFGS iterations for different settings for the two datasets used in \autoref{subsec:faces}.
For these examples, we used the KL loss (since it gave the best reconstructions overall), we did not have to use the log-domain stabilization, and we restarted L-BFGS every 10 iterations.
At an equal number of Sinkhorn iterations ($L$), enabling the warm start always yields better reconstructions after a certain number of iterations.
It comes at a small overhead cost in time (around 25\%) because the L-BFGS line search routine requires more evaluations at the start.
For the example in \autoref{fig:warmstart001}, the computation times are 20 minutes for $L=2$, 25 minutes for the warm restart and $L=2$, and 15 hours for $L=100$.
In this particular case, enabling the warm start with two Sinkhorn iterations yields even better results than having 100 Sinkhorn iterations without a warm start and with a 36 gain factor in time.
For the second dataset (\autoref{fig:warmstart014}), enabling the warm start does not yield results as good as when running 100 Sinkhorn iterations. However, it would require considerably more than two Sinkhorn iterations, and hence a lot more time, to achieve the same result without it.
The computation times in all three cases are similar to the previous example.

\subsection{Sinkhorn heavyball}\label{sec:heavyball}
As part of a generalization of the Sinkhorn algorithm for solving OT between tensor fields~\cite{peyre2016}, Peyré et al. introduced relaxation variables. In the particular case of scalar OT (our framework in the present work), these relaxation variables amount to an averaging step in the Sinkhorn updates; for instance, in the case of the barycenter scaling updates \eqref{eq:a}, \eqref{eq:b},

\begin{align}
\label{eq:atilde}\tilde{a}_s^{(l)} &= \frac{d_s}{Kb_s^{(l-1)}},\\
\nonumber a_s^{(l)} &= \left(a_s^{(l-1)}\right)^\tau \left(\tilde{a}_s^{(l)}\right)^{1-\tau},\\
\label{eq:btilde}\tilde{b}_s^{(l)} &= \frac{P^{(l)}\left(D, \lambda\right)}{K^\top a_s^{(l)}},\\
\nonumber b_s^{(l)} &=  \left(b_s^{(l-1)}\right)^\tau\left(\tilde{b}_s^{(l)}\right)^{1-\tau}.
\end{align}
\noindent $\tau=0$ yields the usual Sinkhorn iterations, but it has been shown that negative values of $\tau$ produce extrapolation and can lead to a considerable increase in the rate of convergence of the Sinkhorn algorithm~\cite[Remark 6]{peyre2016}. This effect can be thought of in the same way as the heavy ball method \cite{nesterov2013, zavriev1993}, often used in optimization problems and dating back to Polyak \cite{polyak1964}, \textit{i.e.} as the addition of a momentum term (\textit{e.g.}, $(a_s^{(l-1)}/\tilde{a}_s^{(l)})^\tau$, which amounts to $\tau(u_s^{(l-1)} - \tilde{u}_s^{(l)})$ in the log-domain) to the usual Sinkhorn updates. This acceleration scheme can be used within our method by applying an automatic differentiation tool~\cite{theano2016} to the forward Sinkhorn loop yielding the barycenter (shown in \autoref{alg:tau} in the Appendix) and feeding the gradients to \autoref{alg:WDL}.

\subsection{Unbalanced}\label{sec:unbal}
In \eqref{eq:transportplans}, we defined the set of admissible transport plans $\Pi(p,q)$ as the set of matrices whose marginals are equal to the two input measures, that is, with rows summing to $p$ and columns summing to $q$. Equivalently, we can reformulate the definition of the approximate Wasserstein distance \eqref{eq:entpen} as

\begin{align*}
W_\gamma(p,q) \eqdef \min_{T\in\mathbb{R}_+^{N\times N}} \langle T, C\rangle + \gamma H(T) + \iota_{\left\{p\right\}}(T\mathds{1}_N) + \iota_{\left\{q\right\}}(T^\top\mathds{1}_N),
\end{align*}
where $\iota$ is the indicator function defined in \eqref{eq:indicator}. Chizat et al. introduce the notion of unbalanced transport problems~\cite{chizat2016}, wherein this equality constraint between the marginals of the OT plan and the input measures is replaced by some other similarity criterion. Using entropic regularization, they introduce matrix scaling algorithms generalizing the Sinkhorn algorithm to compute, among others, unbalanced barycenters. This generalizes the notion of approximate Wasserstein barycenters that we have focused on thus far.

In particular, using the KL divergence between the transport plan's marginals and the input measures allows for less stringent constraints on mass conservation, which can in turn yield barycenters which maintain more of the structure seen in the input measures. This amounts to using the following definition of $W_\gamma$ in the barycenter formulation \eqref{eq:defwassbar}:

\begin{align*}
W_\gamma(p,q) \eqdef \min_{T\in\mathbb{R}_+^{N\times N}} \langle T, C\rangle + \gamma H(T) + \rho\left(\mathrm{KL}(T\mathds{1}_N\vert p) + \mathrm{KL}(T^\top\mathds{1}_N\vert q)\right),
\end{align*}
where $\rho>0$ is the parameter determining how far from the balanced OT case we can stray, with $\rho=\infty$ yielding the usual OT formulation. In this case, the iterative matrix scaling updates \eqref{eq:sinkhornbary}--\eqref{eq:b} read, respectively~\cite{chizat2016}, as

\begin{align*}
P^{(l)}\left(D,\lambda\right) &= \left(\sum_{s=1}^S\lambda_s\left(K^\top a_s^{(l)}\right)^{\frac{\gamma}{\rho+\gamma}}\right)^{\frac{\rho+\gamma}{\gamma}},\\
a_s^{(l)} &= \left(\tilde{a}_s^{(l)}\right)^{\frac{\rho}{\rho+\gamma}}, b_s^{(l)} =  \left(\tilde{b}_s^{(l)}\right)^{\frac{\rho}{\rho+\gamma}},
\end{align*}
where $\tilde{a}_s^{(l)}, \tilde{b}_s^{(l)}$ are obtained from the usual Sinkhorn updates as in \eqref{eq:atilde}, \eqref{eq:btilde}.

\autoref{alg:taurho}, given in the Appendix, performs the barycenter computation (forward loop) including both the unbalanced formulation and the acceleration scheme shown in \autoref{sec:heavyball}. Automatic differentiation can then be performed using an appropriate library~\cite{theano2016} to obtain the dictionary and weights gradients, which can then be plugged into \autoref{alg:WDL} to obtain a representation relying on unbalanced barycenters.

\section{Applications}\label{sec:apps}

\subsection{Comparison with Wasserstein principal geodesics}\label{sec:WPG}
As mentioned in \autoref{sec:PW}, an approach to generalize PCA to the set of probability measures on some space, endowed with the Wasserstein distance, has recently been proposed~\cite{seguy2015}. Given a set of input measures, an approximation of their Wasserstein principal geodesics (WPG) can be computed, namely geodesics that pass through their isobarycenter (in the Wasserstein sense) and are close to all input measures. Because of the close link between Wasserstein geodesics and the Wasserstein barycenter, it would stand to reason that the set of barycenters of $S=2$ atoms learned using our method could be fairly close to the first WPG. In order to test this, and to compare both approaches, we reproduce the setting of the WPG paper~\cite{seguy2015} experiment on the MNIST dataset within our framework.

\begin{figure}[h]
    \centering
    \includegraphics[width=.7\textwidth]{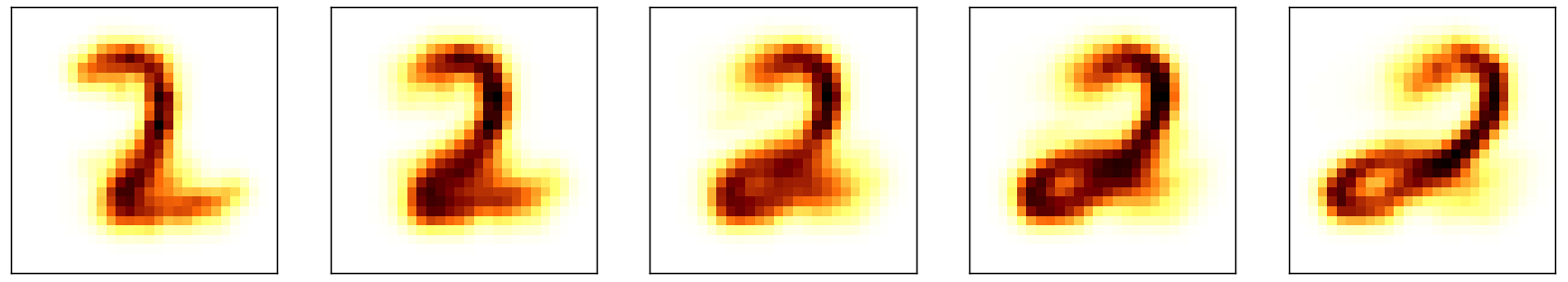}
    \caption{Span of our 2-atom dictionary for weights $(1-t,t), t\in\{0,\frac{1}{4},\frac{1}{2},\frac{3}{4},1\}$, when trained on images of digit $2$.}
    \label{fig:digit2}
\end{figure}
We first run our method to learn two atoms on samples of 1000 images for each of the first four nonzero digits, with parameters $\gamma=2.5,L=30$, and compare the geodesic that runs in between the two learned atoms with the first WPG. An example of the former is shown in \autoref{fig:digit2}. Interestingly, in this case, as with the $3$'s and $4$'s, the two appear visually extremely close (see the first columns of \cite[Figure 5]{seguy2015} for the first WPG). It appears our method \emph{can} thus capture WPGs. We do not seem to recover the first WPG when running on the dataset made up of $1$'s, however. This is not unexpected, as several factors can cause the representation we learn to vary from this computation of the first WPG:
\begin{itemize}
    \item In our case, there is no guarantee the isobarycenter of all input measures lies within the span of the learned dictionary.
    \item Even when it does, since we minimize a non-convex function, the algorithm might converge toward another local minimum.
    \item In this experiment, the WPGs are computed using several approximations~\cite{seguy2015}, including some for the computation of the geodesics themselves, which we are not required to make in order to learn our representation.
\end{itemize}
Note that in the case of this particular experiment (on a subsample of MNIST $1$'s), we tried relaunching our method several times with different random initializations and never observed a span similar to the first WPG computed using these approximations.

\begin{figure}[h]
    \centering
    \includegraphics[width=.4\textwidth]{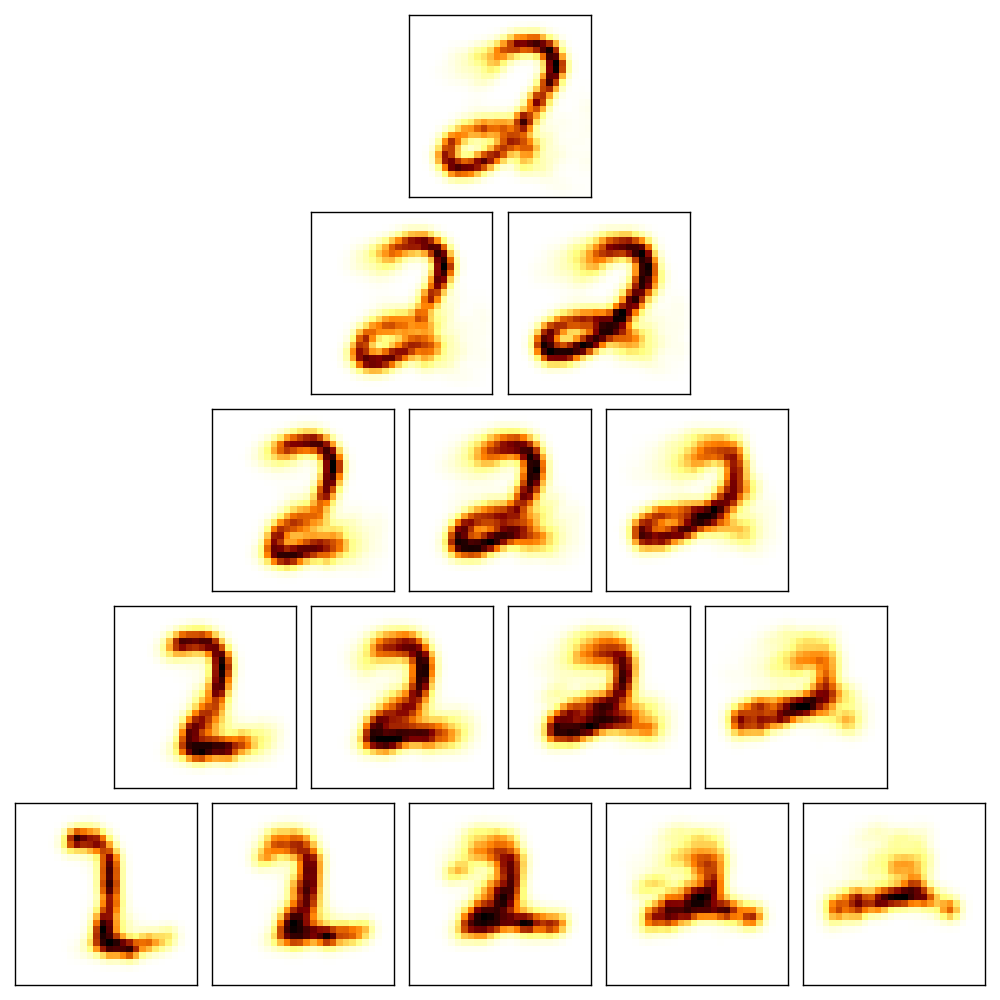}
    \caption{Span of a 3-atom dictionary learned on a set of $2$'s. Weights along each edge are the same as in \autoref{fig:digit2} for the two extreme vertices and $0$ for the other, while the three center barycenters have a weight of $\frac{1}{2}$ for the atom corresponding to the closest vertex and $\frac{1}{4}$ for each of the other two.}
    \label{fig:simplex2}
\end{figure}

Our approach further enables us to combine, in a straightforward way, each of the captured variations when learning more than two atoms. This is illustrated in \autoref{fig:simplex2}, where we run our method with $S=3$. Warpings similar to those captured when learning only $S=2$ atoms (the appearance of a loop within the $2$) are also captured, along with others (shrinking of the vertical size of the digit toward the right). Intermediate values of the weight given to each of the three atoms allow our representation to cover the whole simplex, thus arbitrarily combining any of these captured warpings (\textit{e.g.}, vertically shrinked, loopless $2$ in the middle of the bottom row). 

Figures similar to \autoref{fig:digit2} and \ref{fig:simplex2} for all other digits are given in the Appendix, \autoref{appdx:wpg}.

\subsection{Point spread functions}\label{sec:psf}

\begin{figure}[H]
    \centering
    \begin{subfigure}{.117\linewidth}
        \includegraphics[width=\textwidth]{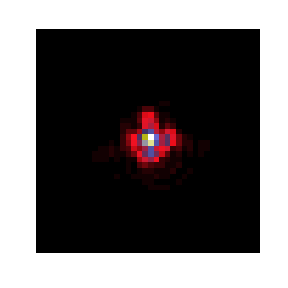}
        \caption{$550$nm}
    \end{subfigure}
    \begin{subfigure}{.117\linewidth}
        \includegraphics[width=\textwidth]{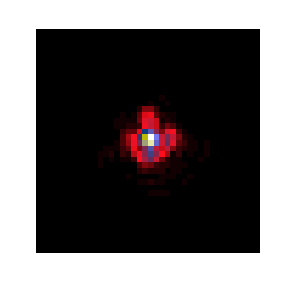}
        \caption{$600$nm}
    \end{subfigure}
    \begin{subfigure}{.117\linewidth}
        \includegraphics[width=\textwidth]{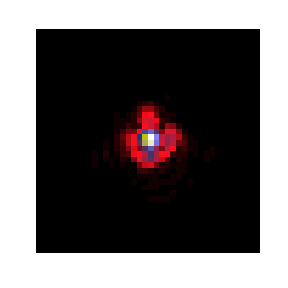}
        \caption{$650$nm}
    \end{subfigure}
    \begin{subfigure}{.117\linewidth}
        \includegraphics[width=\textwidth]{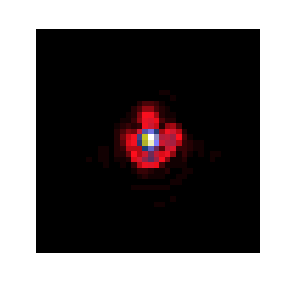}
        \caption{$700$nm}
    \end{subfigure}
    \begin{subfigure}{.117\linewidth}
        \includegraphics[width=\textwidth]{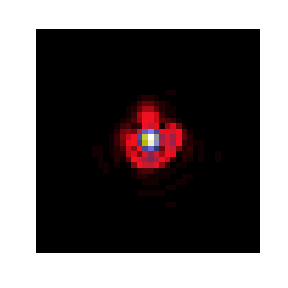}
        \caption{$750$nm}
    \end{subfigure}
    \begin{subfigure}{.117\linewidth}
        \includegraphics[width=\textwidth]{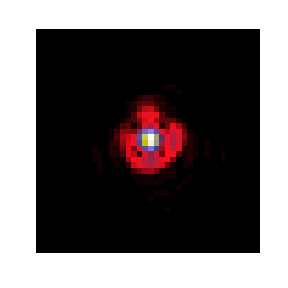}
        \caption{$800$nm}
    \end{subfigure}
    \begin{subfigure}{.117\linewidth}
        \includegraphics[width=\textwidth]{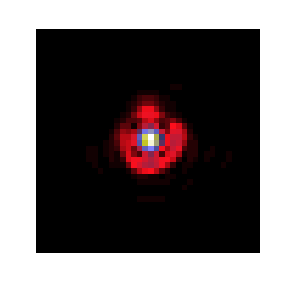}
        \caption{$850$nm}
    \end{subfigure}
    \begin{subfigure}{.117\linewidth}
        \includegraphics[width=\textwidth]{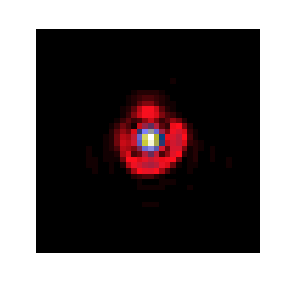}
        \caption{$900$nm}
    \end{subfigure}
    \caption{Simulated Euclid-like PSF variation at a fixed position in the field of view for varying incoming wavelengths.}
    \label{fig:polychrompsf}
    \centering
    \begin{subfigure}{.117\linewidth}
        \includegraphics[width=\textwidth]{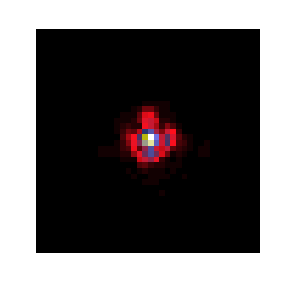}
        \caption{$550$nm}
    \end{subfigure}
    \begin{subfigure}{.117\linewidth}
        \includegraphics[width=\textwidth]{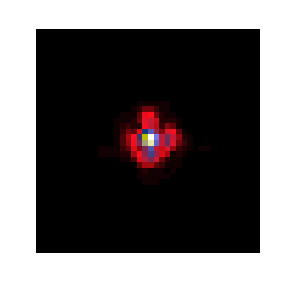}
        \caption{$600$nm}
    \end{subfigure}
    \begin{subfigure}{.117\linewidth}
        \includegraphics[width=\textwidth]{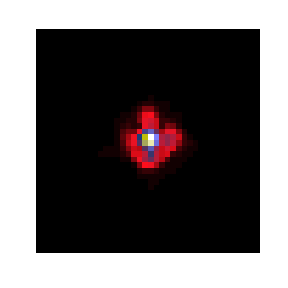}
        \caption{$650$nm}
    \end{subfigure}
    \begin{subfigure}{.117\linewidth}
        \includegraphics[width=\textwidth]{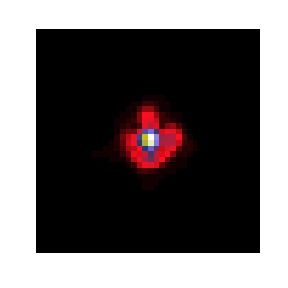}
        \caption{$700$nm}
    \end{subfigure}
    \begin{subfigure}{.117\linewidth}
        \includegraphics[width=\textwidth]{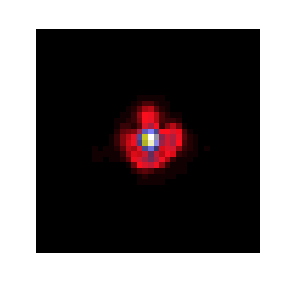}
        \caption{$750$nm}
    \end{subfigure}
    \begin{subfigure}{.117\linewidth}
        \includegraphics[width=\textwidth]{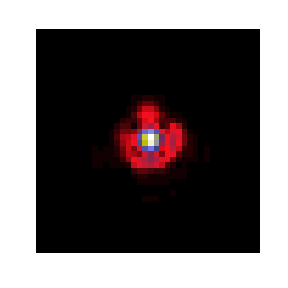}
        \caption{$800$nm}
    \end{subfigure}
    \begin{subfigure}{.117\linewidth}
        \includegraphics[width=\textwidth]{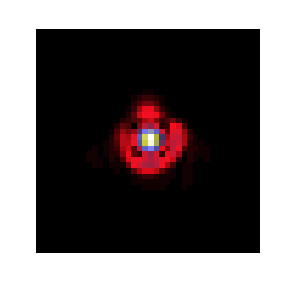}
        \caption{$850$nm}
    \end{subfigure}
    \begin{subfigure}{.117\linewidth}
        \includegraphics[width=\textwidth]{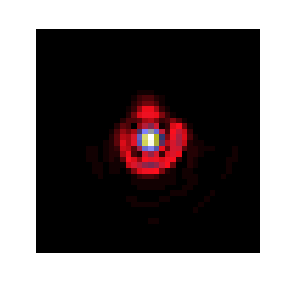}
        \caption{$900$nm}
    \end{subfigure}
    \caption{Polychromatic variations of PSFs by displacement interpolation.}
    \label{fig:WDL_rec}
\end{figure}

\noindent As with every optical system, observations from astrophysical telescopes suffer from a blurring related to the instrument's optics and various other effects (such as the telescope's jitter for space-based instruments). The blurring function, or point spread function (PSF), can vary spatially (across the instrument's field of view), temporally and chromatically (with the incoming light's wavelength). In order to reach its scientific goals, the European Space Agency’s upcoming Euclid space mission~\cite{laureijs2011} will need to measure the shape of one billion galaxies extremely accurately, and therefore correcting the PSF effects is of paramount importance. The use of OT for PSF modeling has been investigated by Irace and Batatia~\cite{irace2013} and Ngolè and Starck~\cite{ngole2017}, both with the aim of capturing the spatial variation of the PSF. For any given position in the field of view, the transformations undergone by the PSF depending on the incoming light's wavelength are also known to contain strong geometrical information, as illustrated in \autoref{fig:polychrompsf}. It is therefore tempting to express these variations as the intermediary steps in the optimal transportation between the PSFs at the two extreme wavelengths. This succession of intermediary steps, the \textit{displacement interpolation} (also known as McCann's interpolation~\cite{mccann1997}) between two measures, can be computed (in the case of the 2-Wasserstein distance) as their Wasserstein barycenters with weights $\lambda = (1-t,t), t\in[0,1]$~\cite{agueh2011}.

We thus ran our method on a dataset of simulated, Euclid-like PSFs~\cite[§4.1]{ngole2015} at various wavelengths and learned only two atoms. The weights were initialized as a projection of the wavelengths into $[0,1]$ but allowed to vary. The atoms were initialized without using any prior information as two uniform images with all pixels set at $1/N$, $N$ being the number of pixels (in this case $40^2$). The fitting loss was quadratic, the entropic parameter $\gamma$ set to a value of $0.5$ to allow for sharp reconstructions, and the number of Sinkhorn iterations set at $120$, with a heavyball parameter $\tau = -0.1$.

\begin{figure}
    \centering
    \begin{subfigure}{0.49\linewidth}
        \includegraphics[width=.7\textwidth]{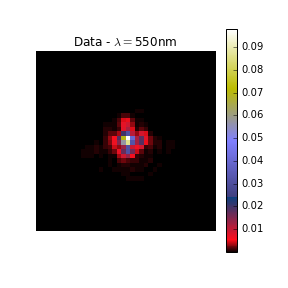}
    \end{subfigure}
    \begin{subfigure}{0.49\linewidth}
        \includegraphics[width=.7\textwidth]{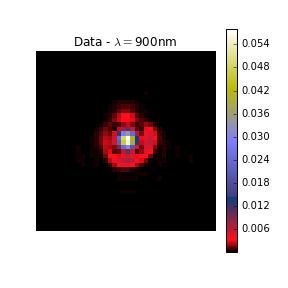}
    \end{subfigure}
    \centering
    \begin{subfigure}{0.49\linewidth}
        \includegraphics[width=.7\textwidth]{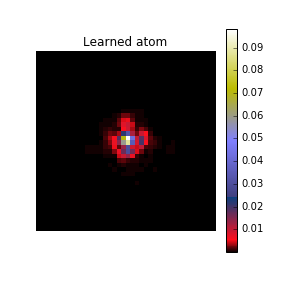}
    \end{subfigure}
    \begin{subfigure}{0.49\linewidth}
        \includegraphics[width=.7\textwidth]{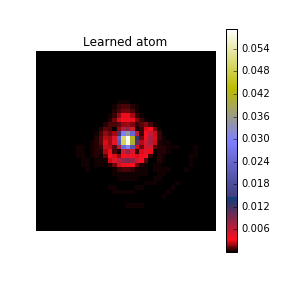}
    \end{subfigure}
    \caption{Extreme wavelength PSFs in the dataset and the atoms making up the learned dictionary.}
    \label{fig:atoms}
\end{figure}

\begin{figure}
    \centering
    \begin{subfigure}{0.49\linewidth}
        \includegraphics[width=.7\textwidth]{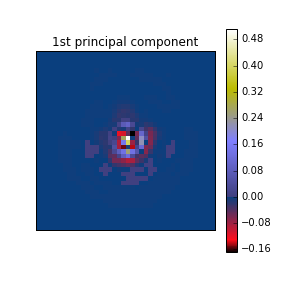}
    \end{subfigure}
    \begin{subfigure}{0.49\linewidth}
        \includegraphics[width=.7\textwidth]{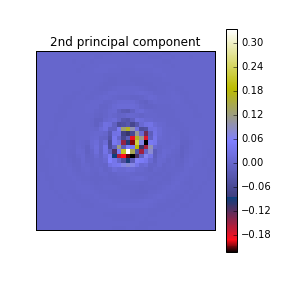}
    \end{subfigure}
    \caption{PCA-learned components.}
    \label{fig:pca_atoms}
\end{figure}

\begin{figure}
    \centering
    \begin{subfigure}{0.49\linewidth}
        \includegraphics[width=\textwidth]{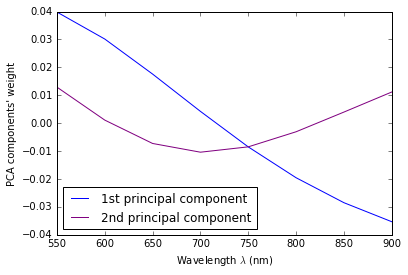}
        \caption{Weights for the first two principal components learned by PCA.}
        \label{fig:pca_wgt}
    \end{subfigure}
    \begin{subfigure}{0.49\linewidth}
        \includegraphics[width=\textwidth]{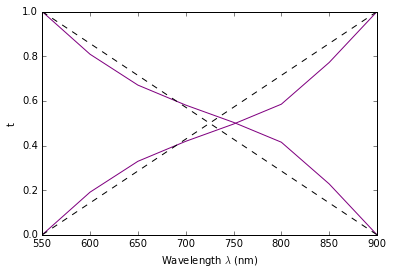}
        \caption{Barycentric weights learned by our method. The dashed lines are the initialization.}
        \label{fig:weights}
    \end{subfigure} 
    \caption{Evolution of representation coefficients by wavelength.}
\end{figure}

The learned atoms, as well as the actual PSFs at both ends of the spectrum, are shown in \autoref{fig:atoms}. Our method does indeed learn atoms that are extremely close visually to the two extremal PSFs. The reconstructed PSFs at the same wavelength as those of \autoref{fig:polychrompsf} are shown in \autoref{fig:WDL_rec} (the corresponding final barycentric weights are shown in \autoref{fig:weights}). This shows that OT, and in particular displacement interpolation, does indeed capture the geometry of the polychromatic transformations undergone by the PSF. On the other hand, when one learns only two components using a PCA, they have no direct interpretation (see \autoref{fig:pca_atoms}), and the weights given to the $2$nd principal component appear to have no direct link to the PSF's wavelength, as shown in \autoref{fig:pca_wgt}.


Note that while adding constraints can also make linear generative methods yield two atoms that are visually close to the extreme PSFs, for instance by using NMF instead of PCA (see \autoref{fig:nmf_atoms}), our method yields lower reconstruction error, with an average normalized mean square error of $1.71\times10^{-3}$ across the whole dataset, as opposed to $2.62\times10^{-3}$ for NMF. As expected, this difference in reconstruction error is particularly noticeable for datapoints corresponding to wavelengths in the middle of the spectrum, as the NMF reconstruction then simply corresponds to a weighted sum of the two atoms, while our method captures more complex warping between them. This shows that the OT representation allows us to better capture the nonlinear geometrical variations due to the optical characteristics of the telescope.

\subsection{Cardiac sequences}\label{sec:cardiac}

We tested our dictionary learning algorithm on a reconstructed MRI sequence of a beating heart.
The goal was to learn a dictionary of four atoms, representing the key frames of the sequence.

\begin{figure}
    \centering
    \includegraphics[width=\textwidth]{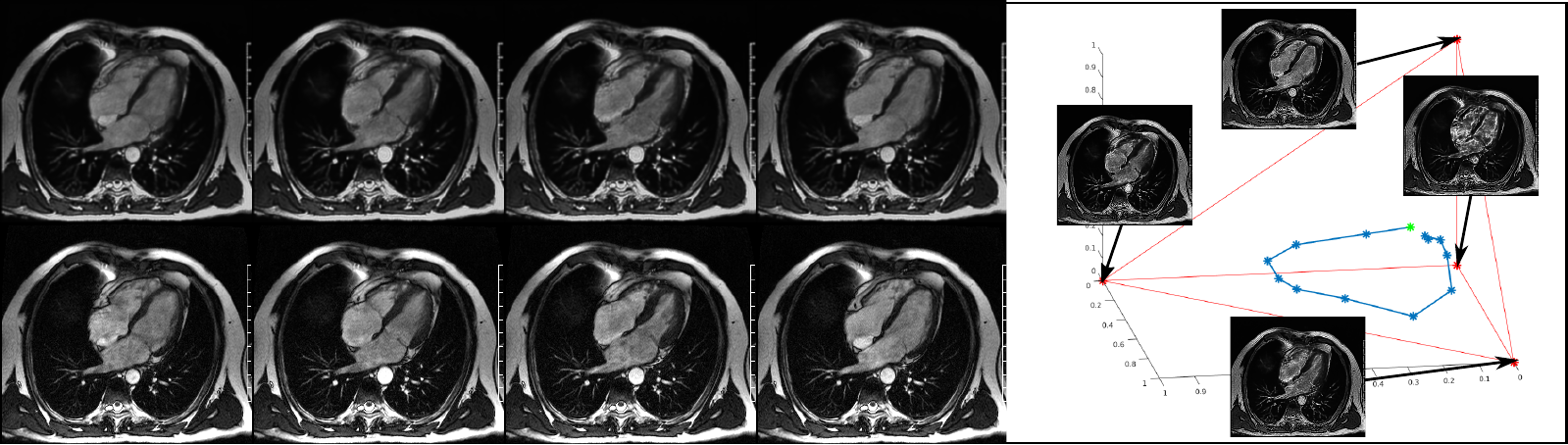}
    \caption{Left: Comparison between fpur frames (out of 13) of the measures (lower row) and the same reconstructed frames (upper row). Right: plot of the reconstructed frames (blue points) by their barycentric coordinates in the 4-atom basis, with each atom (red points) at the vertices of the tetrahedra. The green point is the first frame.}
    \label{fig:heart-single-sequence}
\end{figure}

An advantageous side effect of the weights learned by our method lying in the simplex is that it provides a natural way to visualize them: by associating each atom $d_i$ with a fiducial position $(x_i,y_i)\in\mathbb{R}^2$, each set of weights can be represented as one point placed at the position of the Euclidean barycenter of these positions, with individual weights given to the corresponding atom. Up to rotations and inverse ordering, there are only as many such representations as there are possible orderings of the atoms. In the present case of $S=4$, we can further use the fact that any of the four weights $\lambda_i$ is perfectly known through the other three as $1-\sum_{j\neq i}\lambda_j$. By giving atoms fiducial positions in $\mathbb{R}^3$ and ignoring one of them or, equivalently, assigning it the $(0,0,0)$ position, we thus obtain a unique representation of the weights as seen in \autoref{fig:heart-single-sequence}. The ``barycentric path'' (polyline of the barycentric points) is a cycle, which means the algorithm is successful at finding those key frames that, when interpolated, can represent the whole dataset.
This is confirmed by the similarity between the reconstructions and the input measures.

For this application, we used 13 frames of $272 \times 240$, a regularization $\gamma = 2$, and a scale between weights and atoms of $\zeta =N/(100*M)$, $N = 272\times 240$, $M$ = 13 frames.
Initialization was random for the weights, and constant for the atoms.
We used a quadratic loss because it provided the best results in terms of reconstruction and representation.
We found 25 iterations for the Sinkhorn algorithm to be a good trade-off between computation time and precision.

\subsection{Wasserstein faces}
\label{subsec:faces}

It has been shown that images of faces, when properly aligned, span a low-dimensional space that can be obtained via PCA.
These principal components, called Eigenfaces, have been used for face recognition~\cite{turk1991}. We show that, with the right setting, our dictionary learning algorithm can produce atoms that can be interpreted more easily than their linear counterparts, and can be used to edit a human face's appearance.

We illustrate this application on the MUG facial expression dataset~\cite{aifanti2010}.
From the raw images of the MUG database, we isolated faces and converted the images to grayscale. The resulting images are in \autoref{fig:faces}(a).
We can optionally invert the colors and apply a power factor $\alpha$ similarly to a gamma-correction.
We used a total of 20 ($224\times 224$) images of a single person performing five facial expressions and learned dictionaries of five atoms using PCA, NMF, a K-SVD implementation~\cite{rubinstein2008}, and our proposed method. For the last, we set the number of Sinkhorn iterations to 100 and the maximum number of L-BFGS iterations to 450.
The weights were randomly initialized, and the atoms were initialized as constant.

We performed a cross validation using two datasets, four loss functions, four values for $\alpha$ $(1,2.2,3,5)$, and colors either inverted or not.
We found that none of the $\alpha$ values we tested gave significantly better results (in terms of reconstruction errors).
Interestingly, however, inverting colors improved the result for our method in most cases.
We can conclude that when dealing with faces, it is better to transport the thin and dark zones (eyebrows, mouth, creases) than the large and bright ones (cheeks, forehead, chin).

As illustrated by~\autoref{fig:faces} (and \ref{fig:faces2} in the Appendix), our method reaches similarly successful reconstructions given the low number of atoms, with a slightly higher mean PSNR of 33.8 compared to PSNRs of 33.6, 33.5 and 33.6 for PCA, NMF and K-SVD respectively. 

We show in \autoref{fig:facesLossesB} (and \ref{fig:facesLosses} in Appendix) the atoms obtained when using different loss functions.
This shows how sensible the learned atoms are to the chosen fitting loss, which highlights the necessity for its careful selection if atoms' interpretability is important for the application at hand.

Finally, we showcase an appealing feature of our method: the atoms that it computes allow for facial editing. We demonstrate this application in \autoref{fig:face_editing}. Starting from the isobarycenter of the atoms, by interpolating weights towards a particular atom, we add some of the corresponding emotion to the face.

\newcolumntype{R}[1]{>{\centering\let\newline\\\arraybackslash\begin{sideways}}p{#1}<{\end{sideways}}}

\definecolor{color1}{RGB}{141,160,203}
\definecolor{color2}{RGB}{252,141,98}
\definecolor{color3}{RGB}{102,194,165}
\definecolor{color4}{RGB}{204,102,119}

\newcommand{\colwidth}{0.14\linewidth}
\newcommand{\hspacer}{\hspace{0mm}} 
\newcommand{\CC}[1]{\cellcolor{#1}}
\newcommand{\imcell}[1]{
    \includegraphics[width=\linewidth]{#1}
}

\begin{figure*}
    \centering
    \def\arraystretch{0} 
    \begin{tabular}{p{0.05\linewidth}@{}R{0.05\linewidth}@{\hspacer}>{\centering\let\newline\\\arraybackslash}p{\colwidth}@{\hspacer}>{\centering\let\newline\\\arraybackslash}p{\colwidth}@{\hspacer}>{\centering\let\newline\\\arraybackslash}p{\colwidth}@{\hspacer}>{\centering\let\newline\\\arraybackslash}p{\colwidth}@{\hspacer}>{\centering\let\newline\\\arraybackslash}p{\colwidth}}
        \rowcolor{white}
        &
        (a) Inputs &
        \multicolumn{5}{r}{ \hspace{-1em} \includegraphics[width=0.7\linewidth]{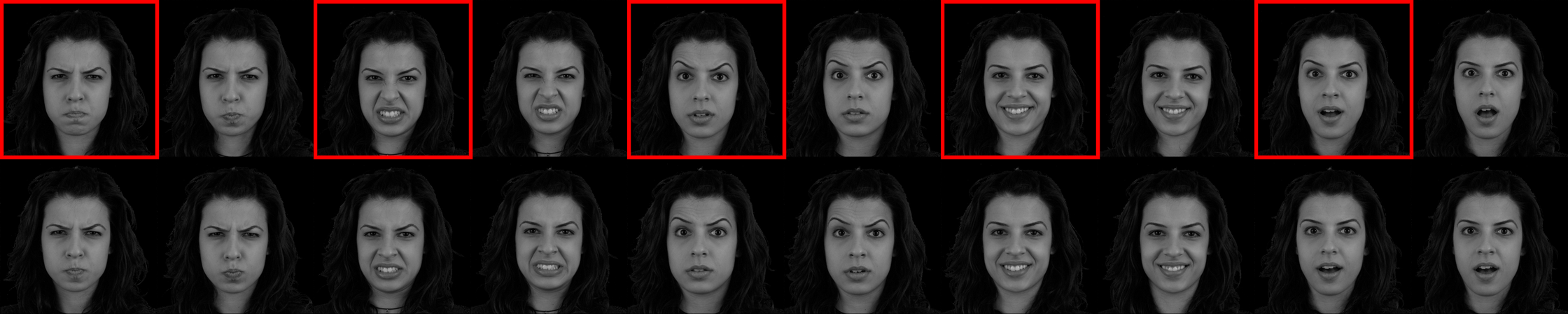}} \\
        \midrule
        
        &
        (b) PCA \CC{color1} &
        \imcell{mug/pc1.png} &
        \imcell{mug/pc2.png} &
        \imcell{mug/pc3.png} &
        \imcell{mug/pc4.png} &
        \imcell{mug/pc5.png} \\
        &
        (c) NMF \CC{color2} &
        \imcell{mug/nmfdic1.png} &
        \imcell{mug/nmfdic2.png} &
        \imcell{mug/nmfdic3.png} &
        \imcell{mug/nmfdic4.png} &
        \imcell{mug/nmfdic5.png} \\
        &
        (d) K-SVD \CC{color3} &
        \imcell{mug/ksvd-dic-1.png} &
        \imcell{mug/ksvd-dic-2.png} &
        \imcell{mug/ksvd-dic-3.png} &
        \imcell{mug/ksvd-dic-4.png} &
        \imcell{mug/ksvd-dic-5.png} \\
        
        \multirow{-4}{*}[30ex]{\rotatebox[origin=c]{90}{Atoms}}
        &
        (e) WDL \CC{color4} &
        \imcell{mug/J0329_KLloss_finalBase_000.png} &
        \imcell{mug/J0329_KLloss_finalBase_001.png} &
        \imcell{mug/J0329_KLloss_finalBase_002.png} &
        \imcell{mug/J0329_KLloss_finalBase_003.png} &
        \imcell{mug/J0329_KLloss_finalBase_004.png} \\
        \midrule
        
        &
        (f) PCA \CC{color1} &
        \imcell{mug/reconsPCA21.png} &
        \imcell{mug/reconsPCA23.png} &
        \imcell{mug/reconsPCA28.png} &
        \imcell{mug/reconsPCA36.png} &
        \imcell{mug/reconsPCA39.png} \\
        &
        (g) NMF \CC{color2} &
        \imcell{mug/reconsNMFDic21.png} &
        \imcell{mug/reconsNMFDic23.png} &
        \imcell{mug/reconsNMFDic28.png} &
        \imcell{mug/reconsNMFDic36.png} &
        \imcell{mug/reconsNMFDic39.png} \\
        &
        (h) K-SVD \CC{color3} &
        \imcell{mug/ksvd-rec-01-21.png} &
        \imcell{mug/ksvd-rec-02-23.png} &
        \imcell{mug/ksvd-rec-04-28.png} &
        \imcell{mug/ksvd-rec-08-36.png} &
        \imcell{mug/ksvd-rec-10-39.png} \\
        \multirow{-4}{*}[35ex]{\begin{sideways}Reconstructions\end{sideways}}
        &
        (i) WDL \CC{color4} &
        \imcell{mug/J0329_KLloss_i-fitting_450_000.png} &
        \imcell{mug/J0329_KLloss_i-fitting_450_001.png} &
        \imcell{mug/J0329_KLloss_i-fitting_450_003.png} &
        \imcell{mug/J0329_KLloss_i-fitting_450_007.png} &
        \imcell{mug/J0329_KLloss_i-fitting_450_009.png} \\
        
    \end{tabular}	
    \caption{We compare our method with Eigenfaces~\protect\cite{turk1991}, NMF and K-SVD~\protect\cite{rubinstein2008} as a tool to represent faces on a low-dimensional space.
        Given a dataset of 20 images of the same person from the MUG dataset~\protect\cite{aifanti2010} performing five facial expressions four times (row (a) illustrates each expression), we project the dataset on the first five Eigenfaces (row (b)).
        The reconstructed faces corresponding to the highlighted input images are shown in row (f).
        Rows (c) and (d), respectively, show atoms obtained using NMF and K-SVD and rows (g) and (h) their respective reconstructions.
        Using our method, we obtain five atoms shown in row (e) that produce the reconstructions in row (i).}
    \label{fig:faces}
\end{figure*}

\begin{figure*}
    \centering
    \begin{tikzpicture}
    \node (table) {
        \def\arraystretch{0} 
        \begin{tabular}{@{}R{0.05\linewidth}@{\hspacer}>{\centering\let\newline\\\arraybackslash}p{0.14\linewidth}@{\hspacer}>{\centering\let\newline\\\arraybackslash}p{0.14\linewidth}@{\hspacer}>{\centering\let\newline\\\arraybackslash}p{0.14\linewidth}@{\hspacer}>{\centering\let\newline\\\arraybackslash}p{0.14\linewidth}@{\hspacer}>{\centering\let\newline\\\arraybackslash}p{0.14\linewidth}@{\hspace{2mm}}>{\centering\let\newline\\\arraybackslash}p{0.14\linewidth}}
        (a) KL loss &
        \imcell{mug/J0373_KLloss_finalBase_000.png} &
        \imcell{mug/J0373_KLloss_finalBase_001.png} &
        \imcell{mug/J0373_KLloss_finalBase_002.png} &
        \imcell{mug/J0373_KLloss_finalBase_003.png} &
        \imcell{mug/J0373_KLloss_finalBase_004.png} &
        \imcell{mug/compareLoss-J0373_KLloss_i-fitting_450_014.png} \\
        (b) Q loss &
        \imcell{mug/J0081_Qloss_finalBase_000.png} &
        \imcell{mug/J0081_Qloss_finalBase_001.png} &
        \imcell{mug/J0081_Qloss_finalBase_002.png} &
        \imcell{mug/J0081_Qloss_finalBase_003.png} &
        \imcell{mug/J0081_Qloss_finalBase_004.png} &
        \imcell{mug/compareLoss-J0081_Qloss_i-fitting_450_014.png} \\
        (c) TV loss &
        \imcell{mug/J0177_TVloss_finalBase_000.png} &
        \imcell{mug/J0177_TVloss_finalBase_001.png} &
        \imcell{mug/J0177_TVloss_finalBase_002.png} &
        \imcell{mug/J0177_TVloss_finalBase_003.png} &
        \imcell{mug/J0177_TVloss_finalBase_004.png} &
        \imcell{mug/compareLoss-J0177_TVloss_i-fitting_450_014.png} \\
        (d) W loss &
        \imcell{mug/J0277_Wloss_finalBase_000.png} &
        \imcell{mug/J0277_Wloss_finalBase_001.png} &
        \imcell{mug/J0277_Wloss_finalBase_002.png} &
        \imcell{mug/J0277_Wloss_finalBase_003.png} &
        \imcell{mug/J0277_Wloss_finalBase_004.png} &
        \imcell{mug/compareLoss-J0277_Wloss_i-fitting_100_014.png} \\
        \end{tabular}
    };
    \draw [red,ultra thick,rounded corners]
    ($(table.north west) !.825! (table.north east)$)
    rectangle
    ($(table.north east) !0.985! ($(table.south west)!.982!(table.south east)$) $);
    \end{tikzpicture}
    
    \caption{We compare the atoms (columns 1 to 5) obtained using different loss functions, ordered by the fidelity of the reconstructions to the input measures (using the mean PSNR), from best to worst: the KL divergence (a) $\overline{PSNR}=32.03$, the quadratic loss (b) $\overline{PSNR}=31.93$, the total variation loss (c) $\overline{PSNR}=31.41$, and the Wasserstein loss (d) $\overline{PSNR}=30.33$. In the last column, we show the reconstruction of the same input image for each loss. We notice that from (a) to (d), the atoms' visual appearance seems to increase even though the reconstruction quality decreases.}
    \label{fig:facesLossesB}
\end{figure*}
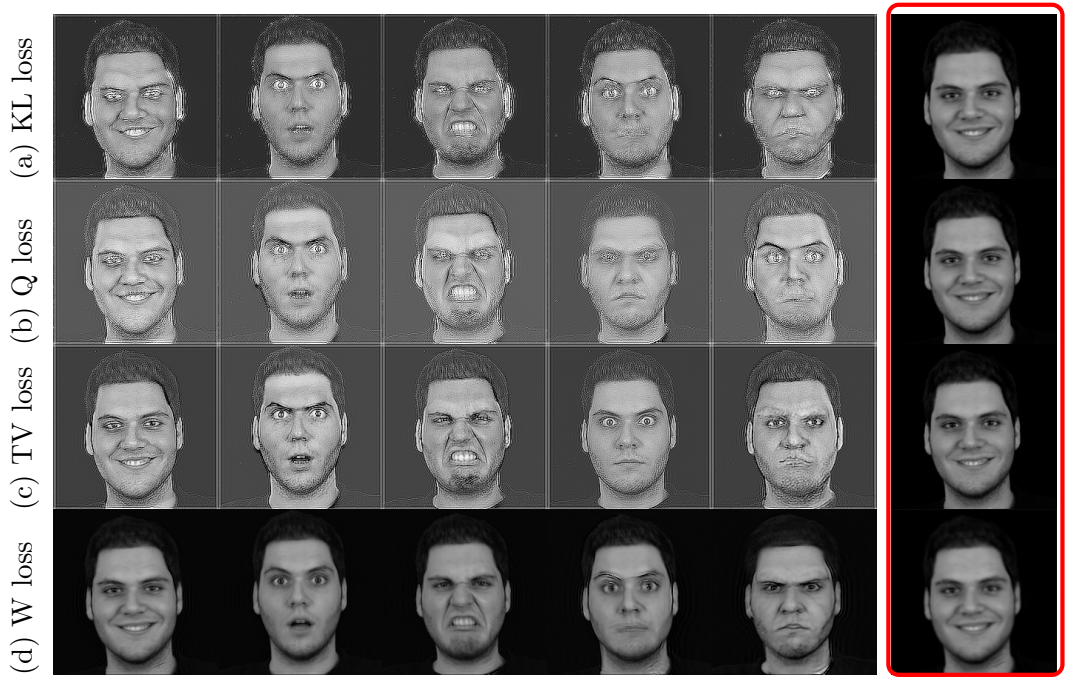

\begin{figure}
    \centering
    \includegraphics[width=.7\textwidth]{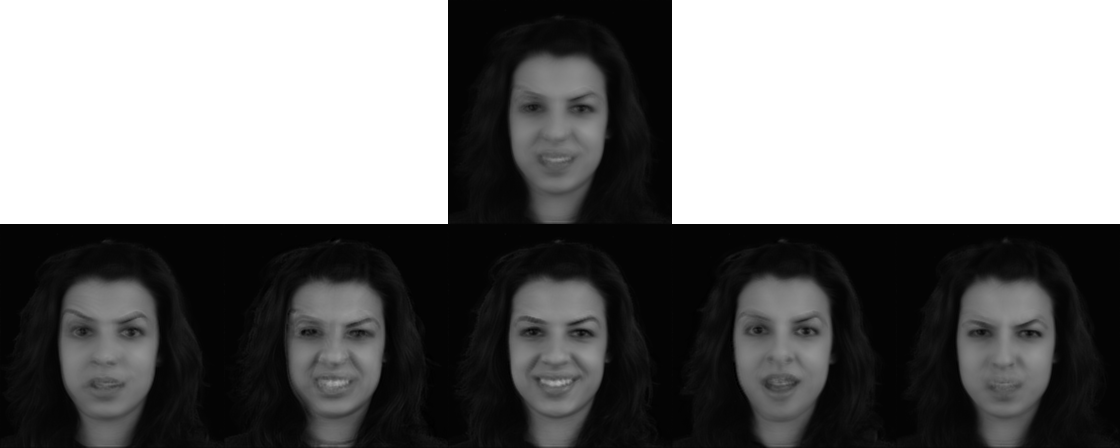}
    \caption{Face editing : Using the atoms shown in row (a) of \protect\autoref{fig:facesLosses}, we interpolate between the atoms' isobarycenter (top image) and each one of the atoms (giving it a relative contribution of 70\%).
        This allows us to emphasize each emotion (bottom images) when starting from a neutral face.}
    \label{fig:face_editing}
\end{figure}

\subsection{Literature learning}

We use our algorithm to represent literary work.
To this end, we use a bag-of-words representation~\cite{salton1986}, where each book is represented by a histogram of its words.
In this particular application, the cost matrix $C$ (distance between each word) is computed exhaustively and stored.
We use a semantic distance between words. These distances were computed from the Euclidian embedding provided by the GloVe database (Global Vectors for Word Representation)~\cite{pennington2014}.

Our learning algorithm is unsupervised and considers similarity between books based on their lexical fields. Consequently we expect it to sort books by either author, writing style, or genre.

To demonstrate our algorithm's performance, we created a database of 20 books by five different authors.
In order to keep the problem size reasonable we only considered words that are between seven and eight letters long.
In our case, it is better to deal with long words because they have a higher chance of holding discriminative information than shorter ones.

\begin{figure}
    \centering
    \includegraphics[width=0.8\textwidth]{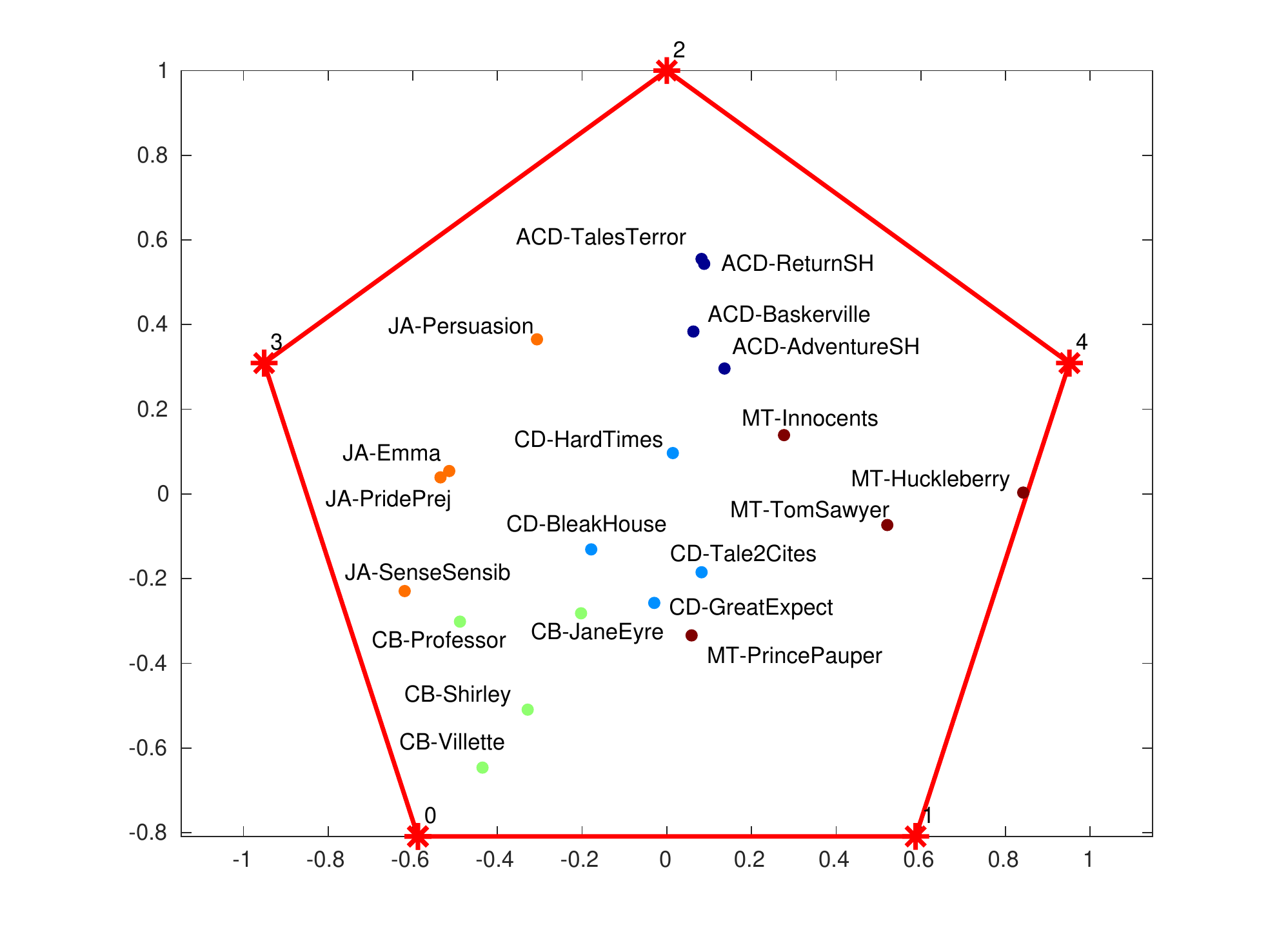}
    \caption{Using our algorithm, we look at word histograms of novels and learn five atoms in a sample of 20 books by five authors.
        Each book is plotted according to its barycentric coordinates with regard to the learned atoms, as explained in \autoref{sec:cardiac}.}
    \label{fig:books5_baryplot}
\end{figure}

The results can be seen in \autoref{fig:books5_baryplot}.
Our algorithm is able to group the novels by author, recognizing the proximity of lexical fields across the different books.
Atom 0 seems to be representing Charlotte Brontë's style, atoms 1 and 4 that of Mark Twain, atom 2 that of Arthur Conan Doyle, and atom 3 that of Jane Austen.
Charles Dickens appears to share an extended amount of vocabulary with the other authors without it differing enough to be represented by its own atom, like others are.

\subsection{Multimodal distributions}
It is a well-known limitation of the regular OT-based Wasserstein barycenters that when there are several distinct areas containing mass, the supports of which are disjoint on the grid, the barycenter operator will still produce barycenters with mass in between them. To illustrate the advantages of using the unbalanced version our method introduced in \autoref{sec:unbal} and the use cases where it might be preferable to do so, we place ourselves in such a setting. 

We generate a dataset as follows: A 1-dimensional grid is separated into three equal parts, and while the center part is left empty, we place two discretized and truncated 1-dimensional Gaussians with the same standard deviation, their mean randomly drawn from every other appropriate position on the grid. We draw 40 such datapoints, yielding several distributions with either one (if the same mean is drawn twice) or two modes in one of the two extreme parts of the grid or one mode in each.

We then run our method in both the balanced and the unbalanced settings. In both cases, $\gamma$ is set to $7$, $100$ Sinkhorn iterations are performed, the loss is quadratic, and the learned dictionary is made up of three atoms. In the unbalanced case, the $\mathrm{KL}$-regularization parameter is set as $\rho = 20$.

\newcommand{\morgancolsizeunbal}{0.221\linewidth}

\begin{figure*}
    \centering
    \begin{tabular}{@{}R{0.05\linewidth}@{\morganspacer}>{\centering\let\newline\\\arraybackslash}p{\morgancolsizeunbal}@{\morganspacer}>{\centering\let\newline\\\arraybackslash}p{\morgancolsizeunbal}@{\morganspacer}>{\centering\let\newline\\\arraybackslash}p{\morgancolsizeunbal}@{\morganspacer}>{\centering\let\newline\\\arraybackslash}p{\morgancolsizeunbal}}
        (a) Balanced & \includegraphics[width=\linewidth]{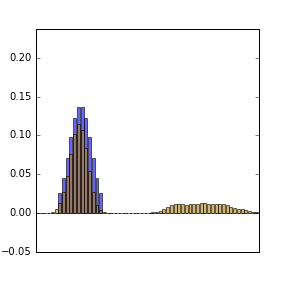} &
        \includegraphics[width=\linewidth]{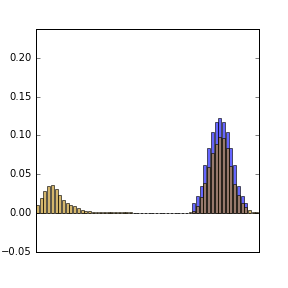} &
        \includegraphics[width=\linewidth]{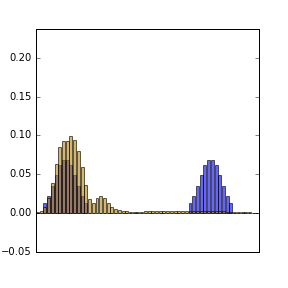} &
        \includegraphics[width=\linewidth]{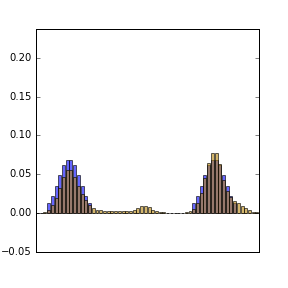}\\
        
        (b) Unbalanced & \includegraphics[width=\linewidth]{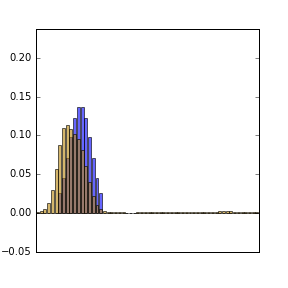} &
        \includegraphics[width=\linewidth]{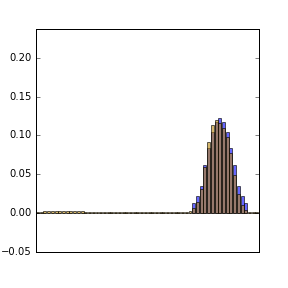} &
        \includegraphics[width=\linewidth]{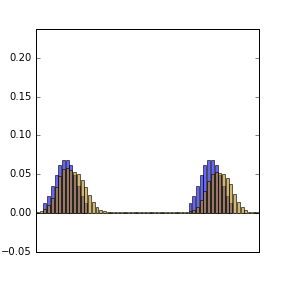} &
        \includegraphics[width=\linewidth]{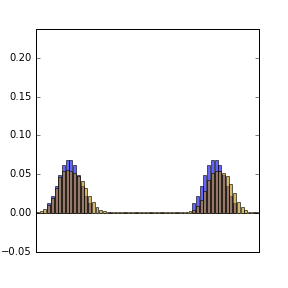}\\

    \end{tabular}	
    \caption{Four different original datapoints (in blue) and their reconstructions (in yellow) from our method in both the balanced (top row) and unbalanced (bottom row) settings. In the balanced case, we see the appearance of spurious modes where there was no mass in the original data or a lack of mass where there was a mode originally (the third example). Conversely, in the unbalanced case, our approach always places mass at the right positions on the grid.}
    \label{fig:unbalex}
\end{figure*}

\autoref{fig:unbalex} shows examples of the input data and its reconstructions in both settings. In the unbalanced case, our method always yields the right number of modes in the right parts of the grid. Running our method with balanced Wasserstein barycenters, however, leads to reconstructions featuring mass in parts of the grid where there was none in the original datapoint (the two left-most examples). Parts of the grid where the datapoint featured a mode can also be reconstructed as empty (the third example). Lastly, we observe mass in areas of the grid that were empty for \emph{all datapoints} (the fourth example).

\section{Conclusion}
This paper introduces a nonlinear dictionary learning approach that uses OT geometry by fitting data to Wasserstein barycenters of a list of learned atoms. We offer schemes to compute this representation based on the addition of an entropic penalty to the definition of OT distances, as well as several variants and extensions of our method. We illustrate the representation our approach yields on several different applications.

Some very recent works present a faster Sinkhorn routine, such as the Greenkhorn algorithm~\cite{altschuler2017} or a multiscale approach~\cite{schmitzer2016}. These could be integrated into our method along with automatic differentiation in order to speed up the algorithm.

\newpage
\appendix

\renewcommand\thefigure{\thesection.\arabic{figure}} 
\setcounter{figure}{0}

\pagebreak
\section{Proof of Proposition 3.1}\label{appdx:proof}
By differentiating \eqref{eq:updateP} with regard to the dictionary or one of the barycentric weights, we can rewrite the Jacobians in \eqref{eq:diffU}, \eqref{eq:difflbda}, respectively, while separating the differentiations with regard to the dictionary $D$, the weights $\lambda_i$ and the scaling vector $b$, by total differentiation and the chain rule:

\begin{align}
\label{eq:diffUP}\left[\partial_D P^{(l)}(D,\lambda)\right]^\top &= \Psi_D^{(l-1)} + B_D^{(l-1)}\Psi_b^{(l-1)},\\
\left[\partial_\lambda P^{(l)}(D,\lambda)\right]^\top &= \Psi_\lambda^{(l-1)} + B_\lambda^{(l-1)}\Psi_b^{(l-1)}.
\end{align}
And, differentiating \eqref{eq:updateb},

\begin{align}
\label{eq:diffBU}B_D^{(l)} &= \Phi_D^{(l-1)} + B_D^{(l-1)}\Phi_b^{(l-1)},\\
B_\lambda^{(l)} &= \Phi_\lambda^{(l-1)} + B_\lambda^{(l-1)}\Phi_b^{(l-1)}.
\end{align}

\noindent We then have, by definitions \eqref{eq:vL}-\eqref{eq:vlm1} and by  plugging \eqref{eq:diffUP} and \eqref{eq:diffBU} into \eqref{eq:diffU},
\begin{align}
\nonumber \nabla_D \mathcal{E}_L(D,\lambda) &= \Psi_D^{(L-1)}\left(\nabla\mathcal{L}(P^{(L)}(D,\lambda),x)\right) + B_D^{(L-1)} v^{(L-1)}\\
\nonumber &= \Psi_D^{(L-1)}\left(\nabla\mathcal{L}(P^{(L)}(D,\lambda),x)\right) + \Phi_D^{(L-2)}\left(v^{(L-1)}\right) + B^{(L-2)}_D \left(v^{(L-2)}\right)\\
\nonumber &= \dots\\
\nabla_D \mathcal{E}_L(D,\lambda)&= \Psi_D^{(L-1)}\left(\nabla\mathcal{L}(P^{(L)}(D,\lambda),x)\right) + \sum_{l=0}^{L-2} \Phi_D^{(l)}\left(v^{(l+1)}\right),
\end{align}
where the sum starts at $0$ because $B_D^{(0)}=0$ since we initialized $b^{(0)}$ as a constant vector. This proves \eqref{eq:nabUE}. Similarly, differentiating with regard to $\lambda$ yields

\begin{align*}
\nabla_\lambda \mathcal{E}_L(D,\lambda)= \Psi_\lambda^{(L-1)}\left(\nabla\mathcal{L}(P^{(L)}(D,\lambda),x)\right) + \sum_{l=0}^{L-2} \Phi_\lambda^{(l)}\left(v^{(l+1)}\right).
\end{align*}
This proves \eqref{eq:nablbdaE}. The detailed derivation of the differentials of $\varphi$, $\Phi$, and $\Psi$ with regard to all three variables is given in the Appendix, \autoref{appdx:derivations}.

\pagebreak
\section{Stabilized backward loop}\label{appdx:logbackward}

\begin{algorithm}
    \begin{program}
        \mbox{\textbf{Inputs:} Data $x\in\Sigma_N$, atoms $d_1,\dots,d_S\in\Sigma_N$, current weights $\lambda\in\Sigma_S$}
        \color{blue}{\COMMENT{Sinkhorn loop}}
        \forall s, v_s^{(0)}  \eqdef \mathbf{0}_{N}\\
        \FOR l=1 \TO L \STEP 1 \DO
        \forall s, \tilde{\varphi}_s^{(l)} \eqdef K_{LS} \left(\log(d_s) - K_{LS}(v_s^{(l-1)})\right)\\
        \tilde{p} \eqdef \sum_s \lambda_s \tilde{\varphi}_s^{(l)}\\
        \forall s, v_s^{(l)} \eqdef \tilde{p}-\tilde{\varphi}_s^{(l)}
        \OD
        p = \exp(\tilde{p})
        \color{blue}{\COMMENT{Backward loop - weights}}
        w \eqdef \mathbf{0}_S\\
        r \eqdef \mathbf{0}_{S\times N}\\
        g \eqdef \nabla \mathcal{L}(p,x) \odot p\\
        \FOR l=L \TO 1 \STEP -1 \DO
        \forall s, w_s \eqdef w_s + \langle\tilde{\varphi}_s^{(l)}, g \rangle\\
        \forall s, \tilde{t}_s \eqdef K_{LS}\left(\log(\lambda_s g - r_s) - \tilde{\varphi}_s^{(l)}\right) + \log(d_s) - 2*K_{LS}(v_s^{(l-1)}) \\
        \forall s, r_s \eqdef \exp\left(K_{LS}(\tilde{t}_s) + v_s^{(l-1)}\right)\\
        g \eqdef -\sum_s r_s
        \OD
        \color{blue}{\COMMENT{Backward loop - dictionary}}
        y \eqdef \mathbf{0}_{S\times N}\\
        z \eqdef \mathbf{0}_{S\times N}\\
        n \eqdef \nabla \mathcal{L}(p,x)
        \FOR l=L \TO 1 \STEP -1 \DO
        \forall s, \tilde{c}_s \eqdef K_{LS}\left(\log(\lambda_sn + z_s) + v_s^{(l)}\right)\\
        \forall s, y_s \eqdef y_s + \exp\left(\tilde{c}_s - K_{LS}(v_s^{(l-1)})\right)\\
        \forall s, z_s \eqdef \exp\left(-\tilde{\varphi}_s^{(l-1)} + K_{LS}\left(\log(d_s) + \tilde{c}_s - 2*K_{LS}(v_s^{(l-1)})\right)\right)\\
        n \eqdef - \sum_s z_s
        \OD
        \mbox{\textbf{Outputs:} $P^{(L)}(D,\lambda) \eqdef p, \nabla_D\mathcal{E}^{(L)} \eqdef y, \nabla_\lambda\mathcal{E}^{(L)} \eqdef w$}
    \end{program}
    \caption{\texttt{logSinkhornGrads}: Computation of dictionary and barycentric weights gradients in log-domain. Log-domain variables are marked with a tilde.}
    \label{alg:logGrads}
\end{algorithm}

\section{Detailed derivations}
\label{appdx:derivations}
Let us first introduce the following notation:
\begin{align}
\varphi\colon \begin{aligned}
\mathbb{R}^{N}\times\mathbb{R}^{N} &\to \mathbb{R}^{N}\\
b_s,d &\mapsto K^\top\frac{d}{Kb_s}\nonumber\end{aligned}.
\end{align}

\subsection{Computation of $\partial_b\varphi$}
By definition,

\begin{align}
\label{eq:diffvarphib}\frac{\partial \varphi}{\partial b_s} (b_s,d) = - K^\top \Delta\left(\frac{d}{(Kb_s)^2}\right) K.
\end{align}
In what follows, we will denote $\varphi_{NS}(b,D) = \left[\varphi(b_1,d_1)^\top,\dots,\varphi(b_S, d_S)^\top\right]^\top \in \mathbb{R}^{NS}$.

\begin{align*}
\partial_b\varphi_{NS}(b,D) = \begin{pmatrix}
\frac{\partial \varphi(b_1,d_1)}{\partial b_1} & \mathbf{0}_{N\times N} & \dots & \mathbf{0}_{N\times N}\\
\mathbf{0}_{N\times N} & \frac{\partial\varphi(b_2,d_2)}{\partial b_2} & \dots & \mathbf{0}_{N\times N}\\
\vdots & & \ddots & \vdots\\
\mathbf{0}_{N\times N} & \dots &\mathbf{0}_{N\times N} & \frac{\partial\varphi(b_S,d_S)}{\partial b_S}
\end{pmatrix}.
\end{align*}

\subsection{Computation of $\Psi_b$}
Taking the logarithm of \eqref{eq:defPsi} yields

\begin{align*}
\log(\Psi(b,D,\lambda)) = \sum_s \lambda_s \log(\varphi(b_s,d_s)),
\end{align*}
the differentiation of which gives us

\begin{align}
\nonumber\Delta\left( \frac{\mathds{1}_N}{\Psi(b,D,\lambda)}\right) \partial_b\Psi(b,D,\lambda) &= \begin{pmatrix}
\lambda_1 I_N & \dots & \lambda_S I_N
\end{pmatrix} \Delta\left(\frac{\mathds{1}_{NS}}{\varphi_{NS}(b,D)}\right) \partial_b \varphi_{NS}(b,D)\\
\label{eq:psib}\implies \Psi_b &= [\partial_b\varphi_{NS}(b,D)]^\top \Delta\left(\frac{\mathds{1}_{NS}}{\varphi_{NS}(b,D)}\right) J_\lambda \Delta(\Psi(b,D,\lambda)),
\end{align}
where $J_\lambda = \begin{pmatrix}
\lambda_1 I_N\\
\vdots\\
\lambda_S I_N
\end{pmatrix} \in \mathbb{R}^{NS\times N}$.

\subsection{Computation of $\Psi_D$}\label{apdx:psiu}
Let $i\in\{1,\dots,S\}.$
\begin{align*}
\Psi(b,D,\lambda) = \prod_{s\neq i} \Delta(\varphi_c(b_s,d_s))^{\lambda_s} . \left(K^\top\frac{d_i}{Kb_i} \right)^{\lambda_i},
\end{align*}
and
\begin{align}
\nonumber\frac{\partial\left(K^\top\frac{d_i}{Kb_i} \right)^{\lambda_i}}{\partial d_i} &= \lambda_i \Delta\left(K^\top\frac{d_i}{Kb_i}\right)^{\lambda_i-1} K^\top \Delta\left(\frac{\mathds{1}_N}{Kb_i}\right)\\
\label{eq:phiunotranspose} \implies \frac{\partial \Psi}{\partial d_i}(b,D,\lambda) &= \lambda_i \frac{\Delta(\Psi(b,D,\lambda))}{\Delta\left(K^\top\frac{d_i}{Kb_i}\right)} K^\top \left(\frac{\mathds{1}_N}{Kb_s}\right).
\end{align}

\subsection{Computation of $\Phi_b$}\label{appdx:phib}
\begin{align}
\nonumber\partial_b\Phi(b,D,\lambda) &= \begin{pmatrix}
\Delta\left(\frac{\mathds{1}_N}{\varphi(b_1,d_1)}\right)\\
\vdots\\
\Delta\left(\frac{\mathds{1}_N}{\varphi(b_S,d_S)}\right)
\end{pmatrix} \partial_b \Psi(b,d) \\
\nonumber&\qquad\qquad-\begin{pmatrix}
\Delta\left(\frac{\Psi(b,D,\lambda)}{\varphi(b_1,d_1)^2}\right)\frac{\partial \varphi(b_1,d_1)}{\partial b_1} & \mathbf{0}_{N\times N} & \dots & \mathbf{0}_{N\times N}\\
\mathbf{0}_{N\times N} & \Delta\left(\frac{\Psi(b,D,\lambda)}{\varphi(b_2,d_2)^2}\right)\frac{\partial\varphi(b_2,d_2)}{\partial b_2} & \dots & \mathbf{0}_{N\times N}\\
\vdots & & \ddots & \vdots\\
\mathbf{0}_{N\times N} & \dots &\mathbf{0}_{N\times N} \nonumber&\Delta\left(\frac{\Psi(b,D,\lambda)}{\varphi(b_S,d_S)^2}\right) \frac{\partial\varphi(b_S,d_S)}{\partial b_S}\
\end{pmatrix}\\
\nonumber&= \Delta\left(\frac{\mathds{1}_{NS}}{\varphi_{NS}(b,D)}\right)I_{N,S}^\top(\partial_b\Psi(b,D,\lambda)) - \Delta\left(\frac{\mathds{1}_{NS}}{\varphi_{NS}(b,D)}\right) \Delta(\Phi(b,D,\lambda))\partial_b\varphi_{NS}(b,D) \\
\nonumber&= \Delta\left(\frac{\mathds{1}_{NS}}{\varphi_{NS}(b,D)}\right) \left[I_{N,S}^\top(\partial_b\Psi(b,D,\lambda)) - \Delta(\Phi(b,D,\lambda))\partial_b\varphi_{NS}(b,D)\right]\\
\nonumber\implies \Phi_b &= \left[\Psi_bI_{N,S} - [\partial_b\varphi_{NS}(b,D)]^\top \Delta(\Phi(b,D,\lambda))\right] \Delta\left(\frac{\mathds{1}_{NS}}{\varphi_{NS}(b,D)}\right)\\
\nonumber&\stackrel{\eqref{eq:psib}}{=} [   [\partial_b\varphi_{NS}(b,D)]^\top \Delta\left(\frac{\mathds{1}_{NS}}{\varphi(b,D)}\right) J_\lambda \Delta(\Psi(b,D,\lambda))I_{N,S} \\
\nonumber&\qquad\qquad-  [\partial_b\varphi_{NS}(b,D)]^\top \Delta(\Phi(b,D,\lambda))] \Delta\left(\frac{\mathds{1}_{NS}}{\varphi_{NS}(b,D)}\right)\\
\label{eq:Phibmiddle}&= [\partial_b\varphi_{NS}(b,D)]^\top \left[  \Delta\left(\frac{\mathds{1}_{NS}}{\varphi(b,D)}\right) J_\lambda \Delta(\Psi(b,D,\lambda))I_{N,S} -   \Delta(\Phi(b,D,\lambda))  \right]
\Delta\left(\frac{\mathds{1}_N}{\varphi_{NS}(b,D)}\right),
\end{align}
where $I_{N,S} = [I_N,\dots,I_N] \in \mathbb{R}^{N\times NS}$. Moreover, we have

\begin{align*}
\Delta\left(\frac{\mathds{1}_{NS}}{\varphi(b,D)}\right) J_\lambda \Delta(\Psi(b,D,\lambda)) &= \begin{pmatrix}
\Delta(1/\varphi(b_1,d_1)) & &\\
& \ddots &\\
&& \Delta(1/\varphi(b_S,d_S))
\end{pmatrix}
\begin{pmatrix}
\lambda_1 \Delta(\Psi(b,D,\lambda))\\
\vdots\\
\lambda_S \Delta(\Psi(b,D,\lambda))
\end{pmatrix}\\
&= \begin{pmatrix}
\lambda_1 \Delta\left(\frac{\Psi(b,D,\lambda)}{\varphi(b_1,d_1)}\right)& &\\
& \ddots &\\
&&\lambda_S \Delta\left(\frac{\Psi(b,D,\lambda)}{\varphi(b_S,d_S)}\right)
\end{pmatrix}\\
&= \Delta(\Phi(b,D,\lambda))\begin{pmatrix}
\lambda_1 I_N\\
\vdots\\
\lambda_S I_N
\end{pmatrix}\\
\Delta\left(\frac{\mathds{1}_{NS}}{\varphi(b,D)}\right) J_\lambda \Delta(\Psi(b,D,\lambda)) &= \Delta(\Phi(b,D,\lambda)) J_\lambda.
\end{align*}
Hence, in \eqref{eq:Phibmiddle},

\begin{align*}
\Phi_b &= [\partial_b\varphi_{NS}(b,D)]^\top \Delta(\Phi(b,D,\lambda)) [J_\lambda I_{N,S} - I_{NS}]
\Delta\left(\frac{\mathds{1}_N}{\varphi_{NS}(b,D)}\right).
\end{align*}

\subsection{Computation of $\Phi_D$}
Let $i\in\{1,\dots\}$. $\forall s\neq i$, the only dependency in $d_i$ of $\Phi^s(b,D,\lambda)$ resides in $\Psi$ (see \eqref{eq:defPhi}), hence

\begin{align*}
\forall s\neq i, \frac{\partial \Phi^s}{\partial d_i} &= \Delta\left(\frac{\mathds{1}_N}{\varphi(b_s,d_s)}\right) \partial_{d_i} \Psi\\
&\stackrel{\eqref{eq:phiunotranspose}}{=} \lambda_i \frac{\Delta(\Psi(B,D,\lambda))}{\Delta(\varphi(b_s,d_s)) \Delta(\varphi(b_i,d_i))} K^\top \Delta\left(\frac{\mathds{1}_N}{Kb_i}\right)\\
&\stackrel{\eqref{eq:defPhi}}{=} \lambda_i \frac{\Delta(\Phi^i(B,D,\lambda))}{\Delta(\varphi(b_s,d_s))} K^\top \Delta\left(\frac{\mathds{1}_N}{Kb_i}\right).
\end{align*}
As for $s=i$, we have

\begin{align*}
\Phi^i(b,D,\lambda) &= \frac{\Psi(b,D,\lambda)}{K^\top\frac{d_i}{Kb_i}}\\
\implies \frac{\partial\Phi^i}{\partial d_i}(b,D,\lambda) &= \Delta\left(\frac{\mathds{1}_N}{\varphi(b_1,d_1)}\right) \partial_D \Psi(b,D,\lambda) - \frac{\Delta(\Psi(b,D,\lambda))}{\Delta(\varphi_i(b_i,d_i)^2)} \partial_{d_i} \varphi(b_i,d_i)\\
&= \Delta\left(\frac{\mathds{1}_N}{\varphi(b_1,d_1)}\right) \partial_D \Psi(b,D,\lambda) -  \frac{\Delta(\Phi^i(b,D,\lambda))}{\Delta(\varphi(b_i,d_i))} K^\top \left(\frac{\mathds{1}_N}{Kb_i}\right)\\
& =(\lambda_i - 1) \frac{\Delta(\Phi^i(b,D,\lambda))}{\Delta(\varphi(b_i,d_i))} K^\top \Delta\left( \frac{\mathds{1}_N}{Kb_i}\right).
\end{align*}

\newpage

\section{Generalized barycenters}

\begin{algorithm}[htpb]
    \begin{program}
        \mbox{\textbf{Inputs:} Data $x\in\Sigma_N$, atoms $d_1,\dots,d_S\in\Sigma_N$, weights $\lambda\in\Sigma_S$,}\\ \mbox{extrapolation parameter $\tau\le 0$}
        \forall s, b_s^{(0)}  \eqdef \mathbf{1}_{N}\\
        \FOR l=1 \TO L \STEP 1 \DO
        \forall s, \tilde{a}_s^{(l)} \eqdef \frac{d_s}{Kb_s^{(l-1)}}\\
        \forall s, a_s^{(l)} \eqdef \left(a_s^{(l-1)}\right)^\tau \left(\tilde{a}_s^{(l)}\right)^{1-\tau}\\
        p \eqdef \prod_s \left(K^\top a_s^{(l)}\right)^{\lambda_s}\\
        \forall s, \tilde{b}_s^{(l)} \eqdef \frac{p}{K^\top a_s^{(l)}}\\
        \forall s, b_s^{(l)} \eqdef  \left(b_s^{(l-1)}\right)^\tau\left(\tilde{b}_s^{(l)}\right)^{1-\tau}\\
        \OD
        \mbox{\textbf{Outputs:} $P^{(L)}(D,\lambda) \eqdef p$}
    \end{program}
    \caption{\texttt{HeavyballSinkhorn}: Computation of approximate Wasserstein barycenters with acceleration}
    \label{alg:tau}
\end{algorithm}

\begin{algorithm}[htpb]
    \begin{program}
        \mbox{\textbf{Inputs:} Data $x\in\Sigma_N$, atoms $d_1,\dots,d_S\in\Sigma_N$, weights $\lambda\in\Sigma_S$,}\\ \mbox{extrapolation parameter $\tau\le 0$, $\mathrm{KL}$ parameter $\rho>0$}
        \forall s, b_s^{(0)}  \eqdef \mathbf{1}_{N}\\
        \FOR l=1 \TO L \STEP 1 \DO
        \forall s, \tilde{a}_s^{(l)} \eqdef \left(\frac{d_s}{Kb_s^{(l-1)}}\right)^{\frac{\rho}{\rho+\gamma}}\\
        \forall s, a_s^{(l)} \eqdef \left(a_s^{(l-1)}\right)^\tau \left(\tilde{a}_s^{(l)}\right)^{1-\tau}\\
        p \eqdef \left(\sum_{s=1}^S\lambda_s\left(K^\top a_s^{(l)}\right)^{\frac{\gamma}{\rho+\gamma}}\right)^{\frac{\rho+\gamma}{\gamma}}\\
        \forall s, \tilde{b}_s^{(l)} \eqdef \left(\frac{p}{K^\top a_s^{(l)}}\right)^{\frac{\rho}{\rho+\gamma}}\\
        \forall s, b_s^{(l)} \eqdef  \left(b_s^{(l-1)}\right)^\tau\left(\tilde{b}_s^{(l)}\right)^{1-\tau}\\
        \OD
        \mbox{\textbf{Outputs:} $P^{(L)}(D,\lambda) \eqdef p$}
    \end{program}
    \caption{\texttt{GeneralizedSinkhorn}: Computation of unbalanced barycenters with acceleration}
    \label{alg:taurho}
\end{algorithm}

\clearpage

\section{Additional results}
\subsection{MNIST and Wasserstein geodesics}\label{appdx:wpg}
This subsection contains additional figures for the application of \autoref{sec:WPG}.
\begin{figure}[H]
    \centering
    \begin{subfigure}{0.24\linewidth}
        \includegraphics[width=\textwidth]{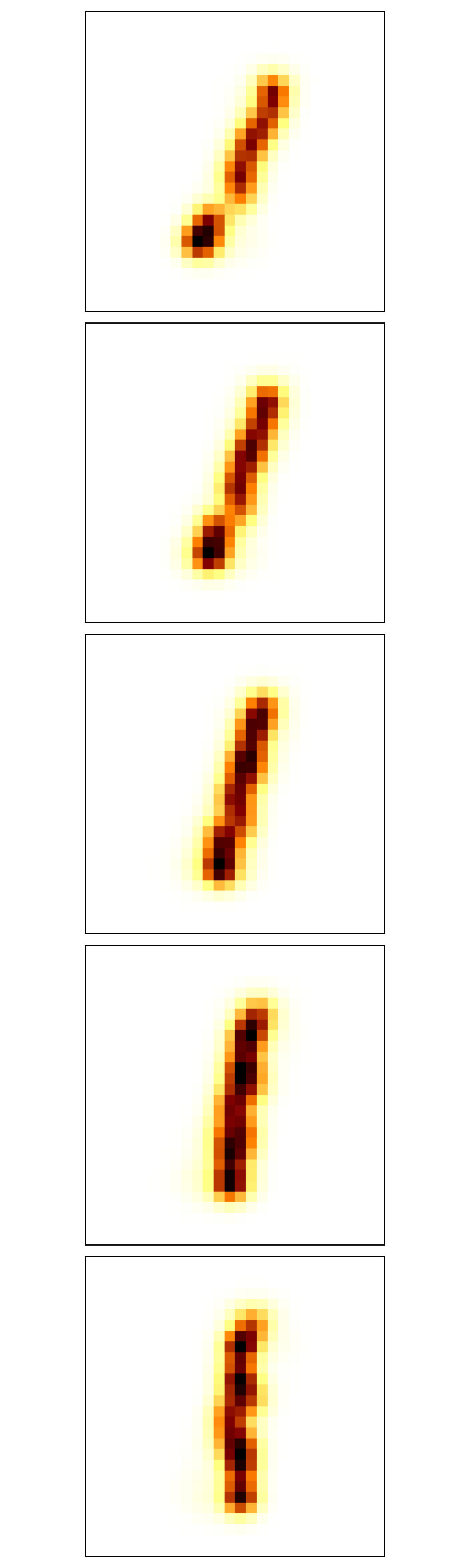}
    \end{subfigure}
    \begin{subfigure}{0.24\linewidth}
        \includegraphics[width=\textwidth]{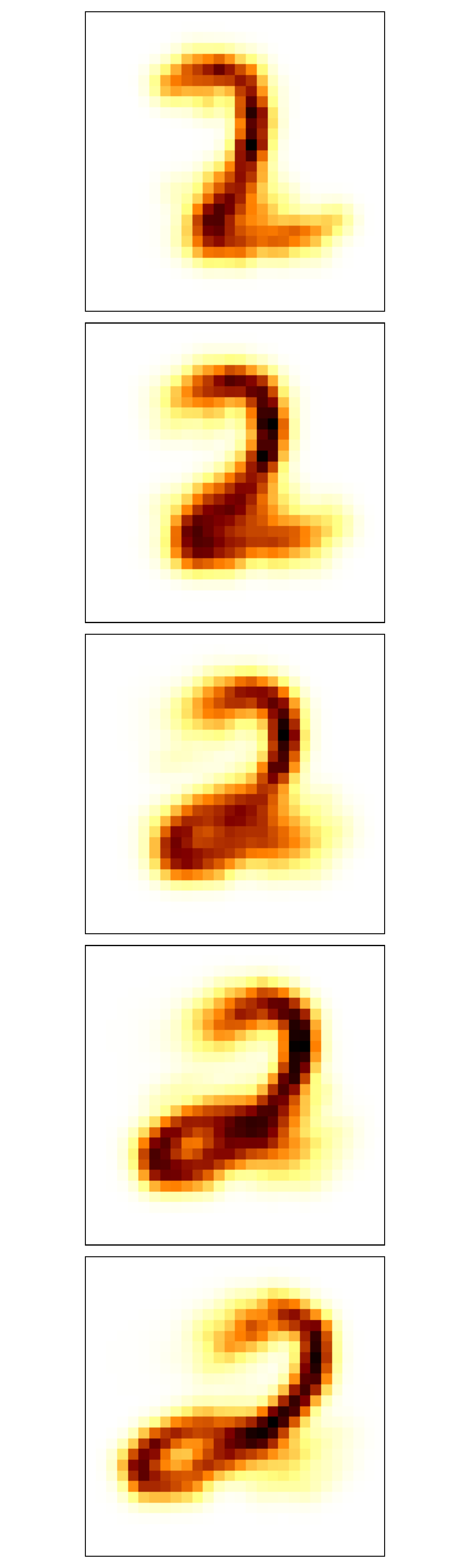}
    \end{subfigure}
    \begin{subfigure}{0.24\linewidth}
        \includegraphics[width=\textwidth]{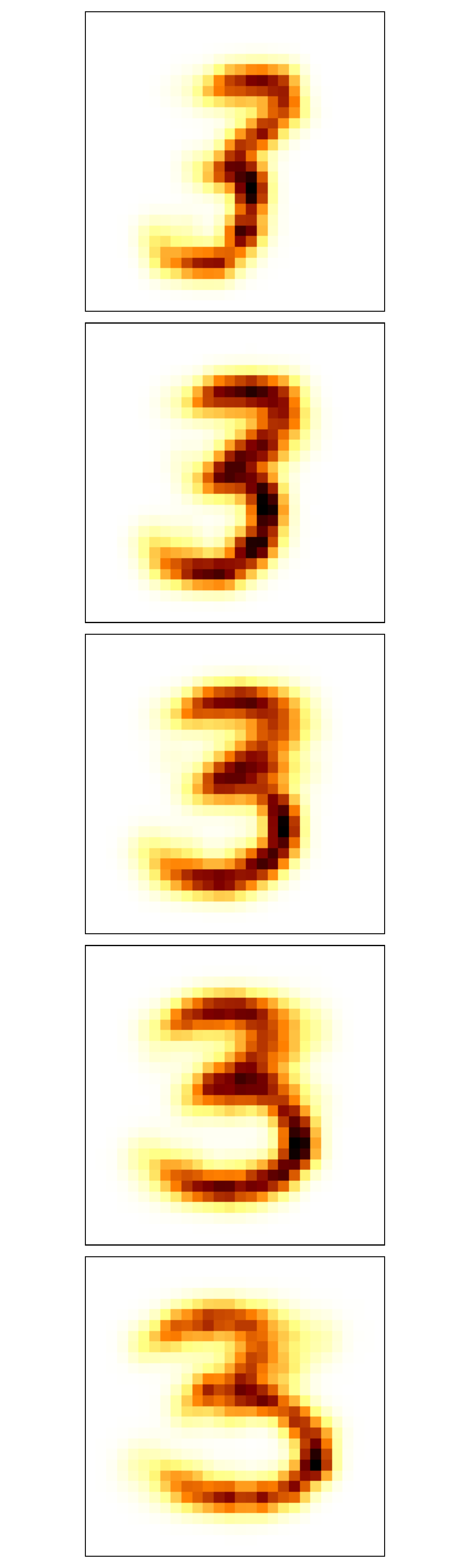}
    \end{subfigure}
    \begin{subfigure}{0.24\linewidth}
        \includegraphics[width=\textwidth]{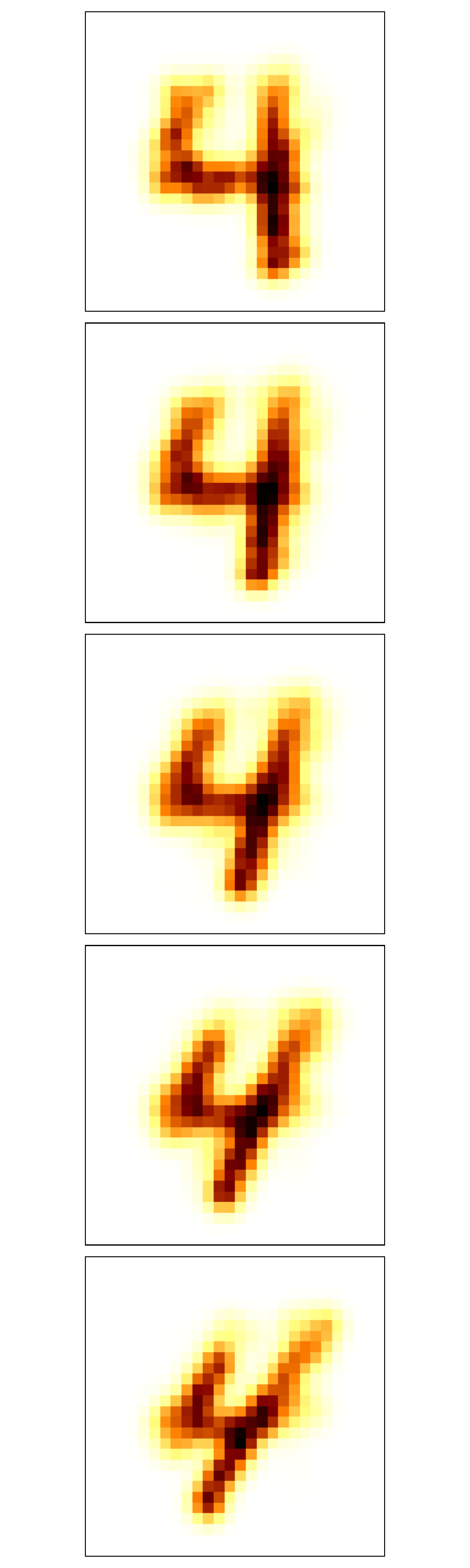}
    \end{subfigure}
    \caption{Span of our two-atom dictionary for weights $(1-t,t), t\in\{0,\frac{1}{4},\frac{1}{2},\frac{3}{4},1\}$ when trained on images of digits $1,2,3,4$. See the first columns of \cite[Figure 5]{seguy2015} for comparison with first WPGs.}
    \label{fig:alldigits}
\end{figure}
\begin{figure}[H]
    \centering
    \includegraphics[width=.4\textwidth]{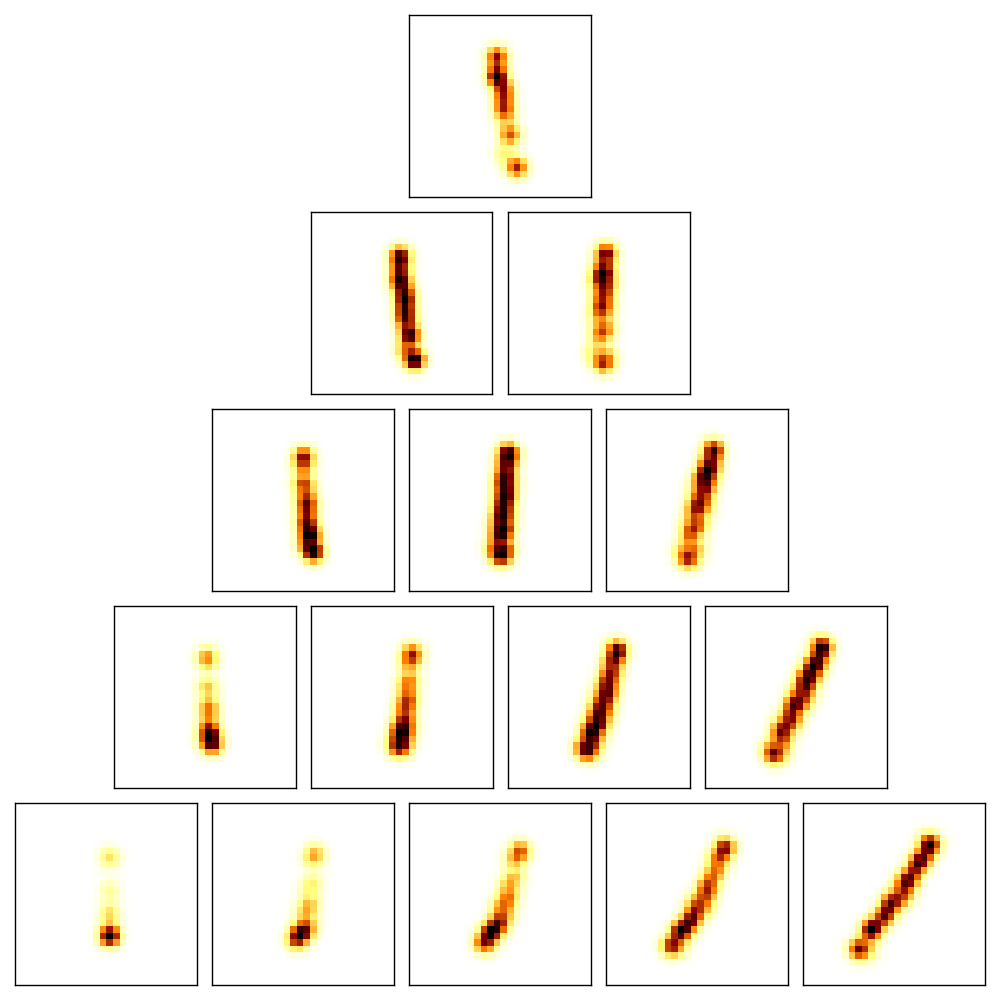}
    \caption{Same as \autoref{fig:simplex2} when training on images of the digit $1$.}
\end{figure}
\begin{figure}[H]
    \centering
    \includegraphics[width=.4\textwidth]{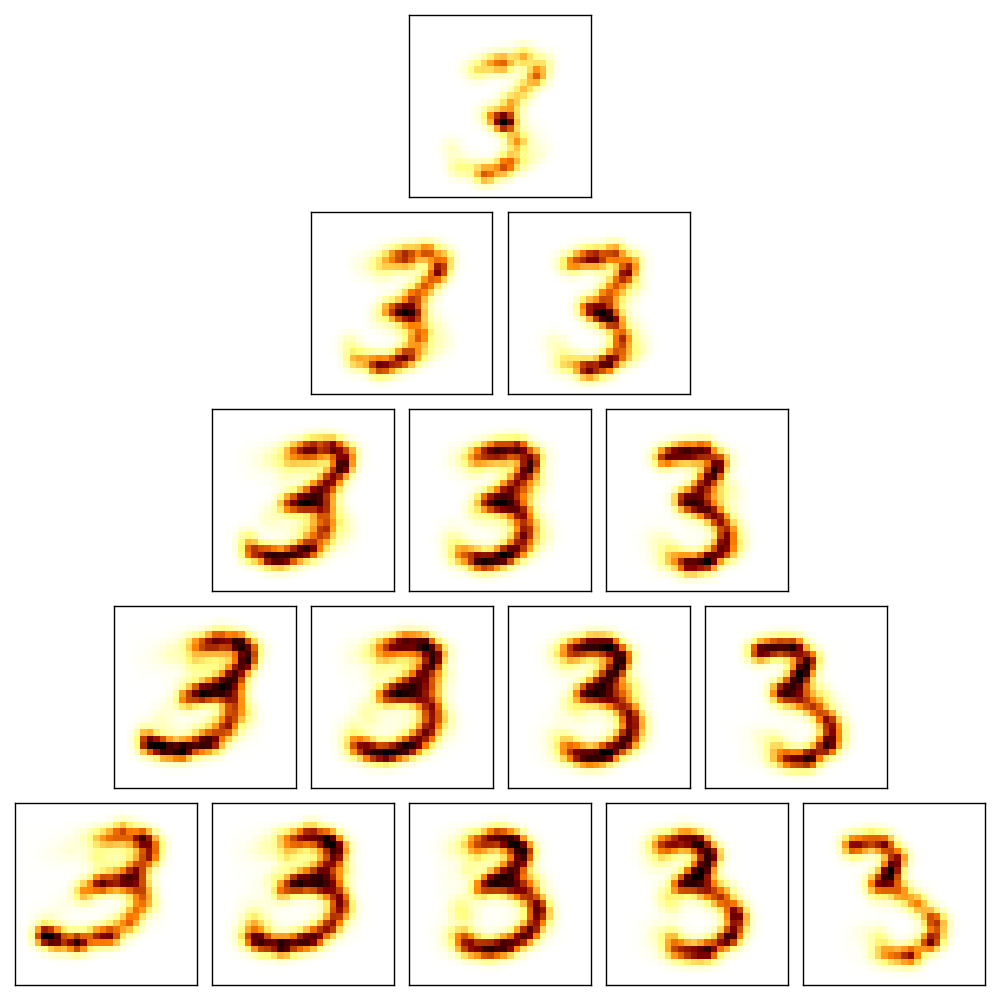}
    \caption{Same as \autoref{fig:simplex2} when training on images of the digit $3$.}
\end{figure}
\begin{figure}[H]
    \centering
    \includegraphics[width=.4\textwidth]{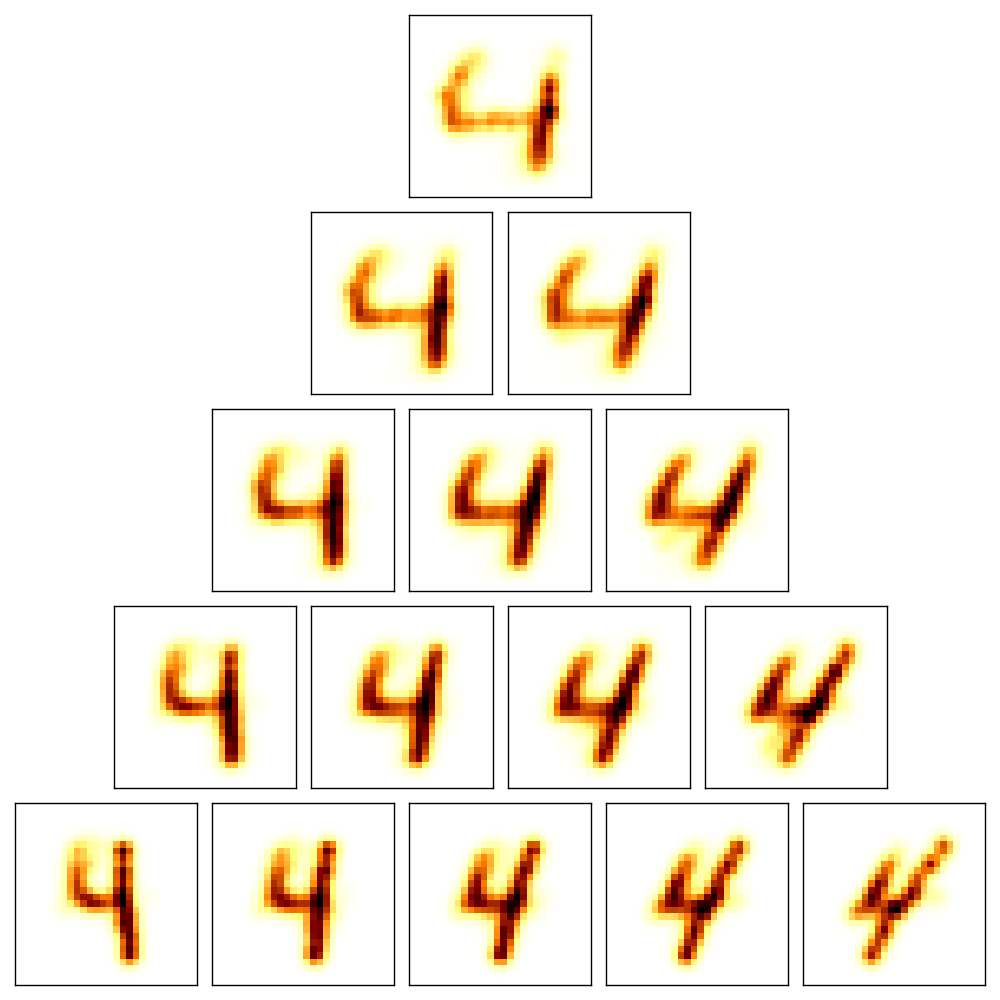}
    \caption{Same as \autoref{fig:simplex2} when training on images of the digit $4$.}
\end{figure}

\subsection{Point spread functions}\label{appdx:psf}
This subsection contains additional figures for the application of \autoref{sec:psf}.
 
\begin{figure}[H]
    \centering
    \begin{subfigure}{0.32\linewidth}
        \includegraphics[width=\textwidth]{apsfpli/dat0.png}
    \end{subfigure}
    \begin{subfigure}{0.32\linewidth}
        \includegraphics[width=\textwidth]{apsfpli/dat1.png}
    \end{subfigure}
    \centering \\
    \vspace{-2em}
    \begin{subfigure}{0.32\linewidth}
        \includegraphics[width=\textwidth]{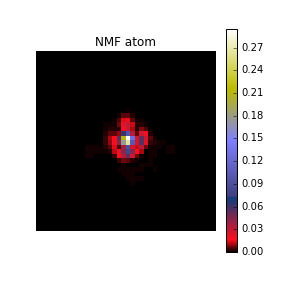}
    \end{subfigure}
    \begin{subfigure}{0.32\linewidth}
        \includegraphics[width=\textwidth]{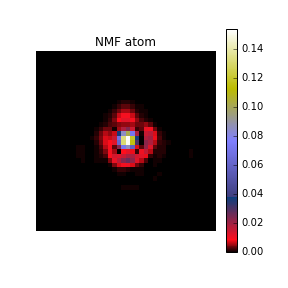}
    \end{subfigure}
    \caption{Extreme wavelength PSFs in the dataset and atoms learned from NMF. See \autoref{fig:atoms} for those learned using our method.}
    \label{fig:nmf_atoms}
\end{figure}

\clearpage
\subsection{Wasserstein faces}
\label{appdx:wassfaces}
This subsection contains additional figures for the application of \autoref{subsec:faces}.

\begin{figure*}[h]
	\centering
	\begin{tikzpicture}
	\node (table) {
		\def\arraystretch{0} 
		\begin{tabular}{@{}R{0.05\linewidth}@{\hspacer}>{\centering\let\newline\\\arraybackslash}p{0.15\linewidth}@{\hspacer}>{\centering\let\newline\\\arraybackslash}p{0.15\linewidth}@{\hspacer}>{\centering\let\newline\\\arraybackslash}p{0.15\linewidth}@{\hspacer}>{\centering\let\newline\\\arraybackslash}p{0.15\linewidth}@{\hspacer}>{\centering\let\newline\\\arraybackslash}p{0.15\linewidth}@{\hspace{2mm}}>{\centering\let\newline\\\arraybackslash}p{0.15\linewidth}}
		(a) KL loss &
		\imcell{mug/J0329_KLloss_finalBase_000.png} &
		\imcell{mug/J0329_KLloss_finalBase_001.png} &
		\imcell{mug/J0329_KLloss_finalBase_002.png} &
		\imcell{mug/J0329_KLloss_finalBase_003.png} &
		\imcell{mug/J0329_KLloss_finalBase_004.png} &
		\imcell{mug/compareLoss-J0329_KLloss_i-fitting_450_018.png} \\
		(b) Q loss &
		\imcell{mug/J0033_Qloss_finalBase_000.png} &
		\imcell{mug/J0033_Qloss_finalBase_001.png} &
		\imcell{mug/J0033_Qloss_finalBase_002.png} &
		\imcell{mug/J0033_Qloss_finalBase_003.png} &
		\imcell{mug/J0033_Qloss_finalBase_004.png} &
		\imcell{mug/compareLoss-J0033_Qloss_i-fitting_450_018.png} \\
		(c) TV loss &
		\imcell{mug/J0133_TVloss_finalBase_000.png} &
		\imcell{mug/J0133_TVloss_finalBase_001.png} &
		\imcell{mug/J0133_TVloss_finalBase_002.png} &
		\imcell{mug/J0133_TVloss_finalBase_003.png} &
		\imcell{mug/J0133_TVloss_finalBase_004.png} &
		\imcell{mug/compareLoss-J0133_TVloss_i-fitting_450_018.png} \\
		(d) W loss &
		\imcell{mug/J0237_Wloss_finalBase_000.png} &
		\imcell{mug/J0237_Wloss_finalBase_001.png} &
		\imcell{mug/J0237_Wloss_finalBase_002.png} &
		\imcell{mug/J0237_Wloss_finalBase_003.png} &
		\imcell{mug/J0237_Wloss_finalBase_004.png} &
		\imcell{mug/compareLoss-J0237_Wloss_i-fitting_150_018.png} \\
		\end{tabular}
	};
	\draw [red,ultra thick,rounded corners]
	($(table.north west) !.825! (table.north east)$)
	rectangle 
	($(table.north east) !0.985! ($(table.south west) !.982! (table.south east)$)$);
	\end{tikzpicture}
	
	\caption{Similarly to~\protect\autoref{fig:facesLossesB}, we compare the atoms obtained using different loss functions, ranking them by mean PSNR: (a) $\overline{PSNR}=33.81$, (b) $\overline{PSNR}=33.72$, (c) $\overline{PSNR}=32.95$, and (d) $\overline{PSNR}=32.34$.}
	\label{fig:facesLosses}
\end{figure*}
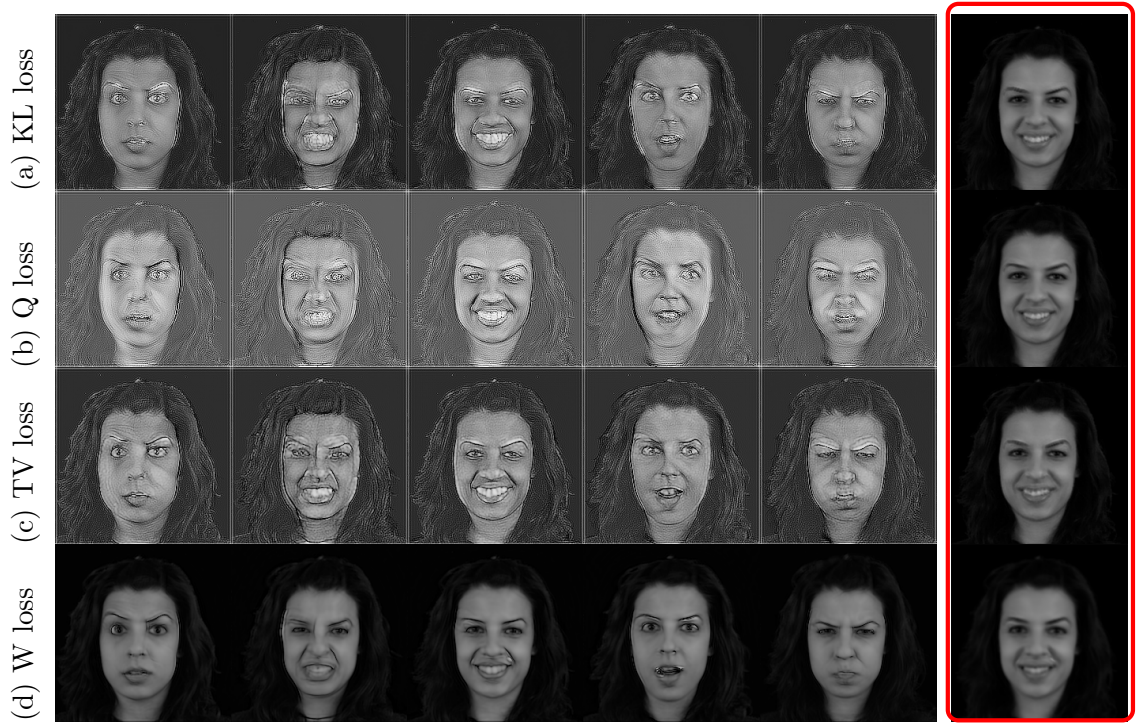

\newcolumntype{R}[1]{>{\centering\let\newline\\\arraybackslash\begin{sideways}}p{#1}<{\end{sideways}}}

\definecolor{color1}{RGB}{141,160,203}
\definecolor{color2}{RGB}{252,141,98}
\definecolor{color3}{RGB}{102,194,165}
\definecolor{color4}{RGB}{204,102,119}

\begin{figure*}[b!]
    \centering
    \def\arraystretch{0}
    \begin{tabular}{p{0.05\linewidth}@{}R{0.05\linewidth}@{\hspacer}>{\centering\let\newline\\\arraybackslash}p{\colwidth}@{\hspacer}>{\centering\let\newline\\\arraybackslash}p{\colwidth}@{\hspacer}>{\centering\let\newline\\\arraybackslash}p{\colwidth}@{\hspacer}>{\centering\let\newline\\\arraybackslash}p{\colwidth}@{\hspacer}>{\centering\let\newline\\\arraybackslash}p{\colwidth}}
        &
        (a) Inputs & \multicolumn{5}{r}{\hspace{-1em}\includegraphics[width=0.8\linewidth]{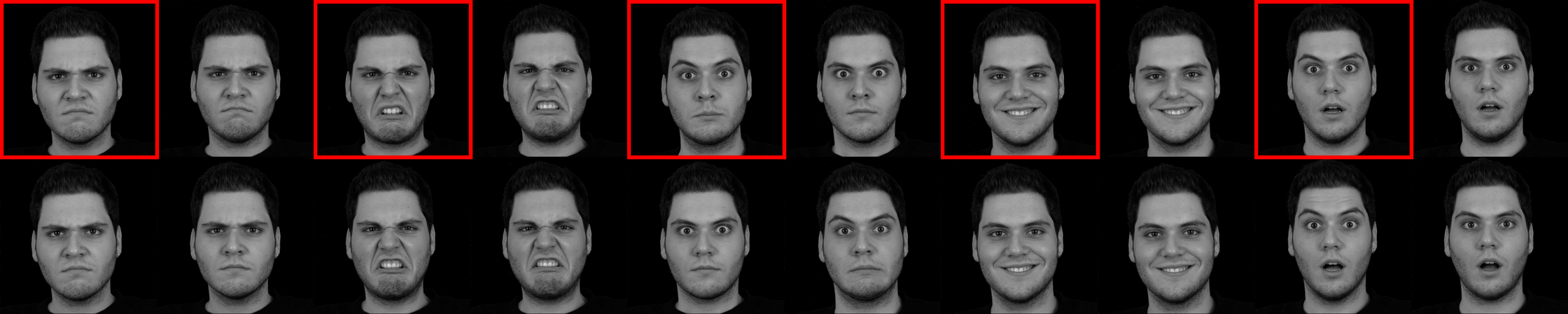}} \\
        \midrule
        &
        (b) PCA \CC{color1} &
        \imcell{mug/pcB1.png} &
        \imcell{mug/pcB2.png} &
        \imcell{mug/pcB3.png} &
        \imcell{mug/pcB4.png} &
        \imcell{mug/pcB5.png} \\
        &
        (c) NMF \CC{color2} &
        \imcell{mug/nmfdicB1.png} &
        \imcell{mug/nmfdicB2.png} &
        \imcell{mug/nmfdicB3.png} &
        \imcell{mug/nmfdicB4.png} &
        \imcell{mug/nmfdicB5.png} \\
        &
        (d) K-SVD \CC{color3} &
        \imcell{mug/ksvd-dic-B1.png} &
        \imcell{mug/ksvd-dic-B2.png} &
        \imcell{mug/ksvd-dic-B3.png} &
        \imcell{mug/ksvd-dic-B4.png} &
        \imcell{mug/ksvd-dic-B5.png} \\
        
        \multirow{-4}{*}[30ex]{\rotatebox[origin=c]{90}{Atoms}}
        &
        (e) WDL \CC{color4} &
        \imcell{mug/J0373_KLloss_finalBase_000.png} &
        \imcell{mug/J0373_KLloss_finalBase_001.png} &
        \imcell{mug/J0373_KLloss_finalBase_002.png} &
        \imcell{mug/J0373_KLloss_finalBase_003.png} &
        \imcell{mug/J0373_KLloss_finalBase_004.png} \\
        \midrule
        &
        (f) PCA \CC{color1} &
        \imcell{mug/reconsPCAB39.png} &
        \imcell{mug/reconsPCAB43.png} &
        \imcell{mug/reconsPCAB110.png} &
        \imcell{mug/reconsPCAB40.png} &
        \imcell{mug/reconsPCAB23.png} \\
        &
        (g) NMF \CC{color2} &
        \imcell{mug/reconsNMFDicB39.png} &
        \imcell{mug/reconsNMFDicB43.png} &
        \imcell{mug/reconsNMFDicB110.png} &
        \imcell{mug/reconsNMFDicB40.png} &
        \imcell{mug/reconsNMFDicB23.png} \\
        &
        (h) K-SVD \CC{color3} &
        \imcell{mug/ksvd-rec-B07-39.png} &
        \imcell{mug/ksvd-rec-B11-43.png} &
        \imcell{mug/ksvd-rec-B19-110.png} &
        \imcell{mug/ksvd-rec-B08-40.png} &
        \imcell{mug/ksvd-rec-B02-23.png} \\
        
        \multirow{-4}{*}[35ex]{\begin{sideways}Reconstructions\end{sideways}}
        &
        (i) WDL \CC{color4} &
        \imcell{mug/J0373_KLloss_i-fitting_450_006.png} &
        \imcell{mug/J0373_KLloss_i-fitting_450_010.png} &
        \imcell{mug/J0373_KLloss_i-fitting_450_018.png} &
        \imcell{mug/J0373_KLloss_i-fitting_450_007.png} &
        \imcell{mug/J0373_KLloss_i-fitting_450_001.png} \\
        
    \end{tabular}
    \caption{Similarly to~\protect\autoref{fig:faces}, we compare our method to the Eigenfaces~\protect\cite{turk1991} approach, NMF and K-SVD as a tool to represent faces on a low-dimensional space.}
    \label{fig:faces2}
\end{figure*}

\clearpage

\section*{Acknowledgements}
This work was supported by the Centre National d'Etudes Spatiales (CNES), the European Community through the grants DEDALE (contract no. 665044) and NORIA (contract no. 724175) within the H2020 Framework Program and the French National Research Agency (ANR) through the grant ROOT. The authors are grateful to the anonymous referees for their salient comments and suggestions, as well as to the SIAM editorial team for their many suggested corrections.

\bibliographystyle{siam}
\bibliography{preprint}
\end{document}